\documentclass[runningheads]{llncs}
\usepackage{graphicx}
\usepackage{amsmath,amssymb} 
\usepackage{color}

\usepackage{subcaption}
\usepackage{url}
\usepackage{mathrsfs}
\usepackage{booktabs}
\usepackage{multirow}
\usepackage{array}
\usepackage{csquotes}
\usepackage{float}

\newcommand{\figref}[1]{Fig.~\ref{#1}}

\newcommand{\tabref}[1]{Table~\ref{#1}}

\newcommand{\equref}[1]{Eq.~\ref{#1}}

\newcommand{\secref}[1]{Sec.~\ref{#1}}

\newcommand{\etal}{\textit{et al}.~}

\newcommand{\eg}{\textit{e}.\textit{g}.~}

\begin{document}
\pagestyle{headings}
\mainmatter
\def\ECCV18SubNumber{***}  

\title{Multichannel Semantic Segmentation \\ with Unsupervised Domain Adaptation}

\titlerunning{Mulitichannel Semantic Segmentation with UDA}

\authorrunning{K.Watanabe, K.Saito, Y.Ushiku, and T.Harada}

\author{Kohei Watanabe\inst{1} \and
Kuniaki Saito\inst{1} \and
Yoshitaka Ushiku \inst{1} \and
Tatsuya Harada \inst{1,2}}

\institute{The University of Tokyo \\
\email{\{watanabe, k-saito, ushiku, harada\}@mi.t.u-tokyo.ac.jp}
\and RIKEN}

\maketitle

\begin{abstract}

Most contemporary robots have depth sensors, and research on semantic segmentation with RGBD images has shown that depth images boost the accuracy of segmentation.
Since it is time-consuming to annotate images with semantic labels per pixel, it would be ideal if we could avoid this laborious work by utilizing an existing dataset or a synthetic dataset which we can generate on our own.
Robot motions are often tested in a synthetic environment, where multichannel (\eg, RGB + depth + instance boundary) images plus their pixel-level semantic labels are available.
However, models trained simply on synthetic images tend to demonstrate poor performance on real images.
In order to address this, we propose two approaches that can efficiently exploit multichannel inputs combined with an unsupervised domain adaptation (UDA) algorithm.
One is a fusion-based approach that uses depth images as inputs.
The other is a multitask learning approach that uses depth images as outputs.
We demonstrated that the segmentation results were improved by using a multitask learning approach with a post-process and created a benchmark for this task.

\end{abstract}

\section{Introduction}
Semantic segmentation is a fundamental task for robots to understand their surroundings in detail.
Most robots have depth sensors. In fact, research on semantic segmentation with RGBD images has been conducted and has demonstrated that depth images boost the accuracy of segmentation~\cite{FuseNet}.
However, semantic pixel-level labels are necessary to train semantic segmentation models in general and it is time-consuming to annotate the image per pixel.
For instance, the pixel labeling of one Cityscapes image takes 1.5 hours on average~\cite{CityScapes}.
It would be ideal to avoid this laborious work by utilizing an existing dataset or a synthetic dataset which we could generate on our own.

Recently, the number of RGBD datasets taken in the real world has increased.
In addition to the widely used 2.5D dataset such as NYUDv2~\cite{silberman2012nyu} and SUNRGBD~\cite{SunRGBD}, large-scale real 3D datasets such as Stanford2d3d~\cite{armeni_cvpr16,armeni2017joint}, ScanNet~\cite{dai2017scannet} have been generated due to the development of the 3D scanner and scalable RGB-D capture system.
However, the number of real datasets is small compared to the RGB dataset.

Conversely, computer graphics technology has been developed and large-scale synthetic datasets have also been generated.
For example, SUNCG~\cite{zhang2017physically} contains 400K physically-based rendered images from 45K realistic 3D indoor scenes.
SceneNet~\cite{McCormac:etal:ICCV2017} contains 5 million images rendered of 16,895 indoor scenes.
It is also possible to purchase 3D CAD models online and create customized synthetic datasets using UnrealCV~\cite{qiu2017unrealcv}.
The appearance of synthetic images is a bit different from that of real ones but the synthetic datasets still look real. In fact, it is ideal if a model trained on these dataset performs well on real datasets because robot motions are often tested in a synthetic environment before being tested in a real environment.

However, such a model is known not to generalize well because of the pixel-level distribution shift~\cite{hoffman2016fcns}.
In order to solve this problem, a domain adaptation technique is necessary.
Although several research studies on unsupervised domain adaptation for semantic segmentation have been conducted, they use only RGB input and do not consider the utilization of a multichannel (here we mean RGB + depth images, which are now easy to obtain in both synthetic and real environments.

\begin{figure}[t]
  \centering
\includegraphics[width=0.8\hsize]{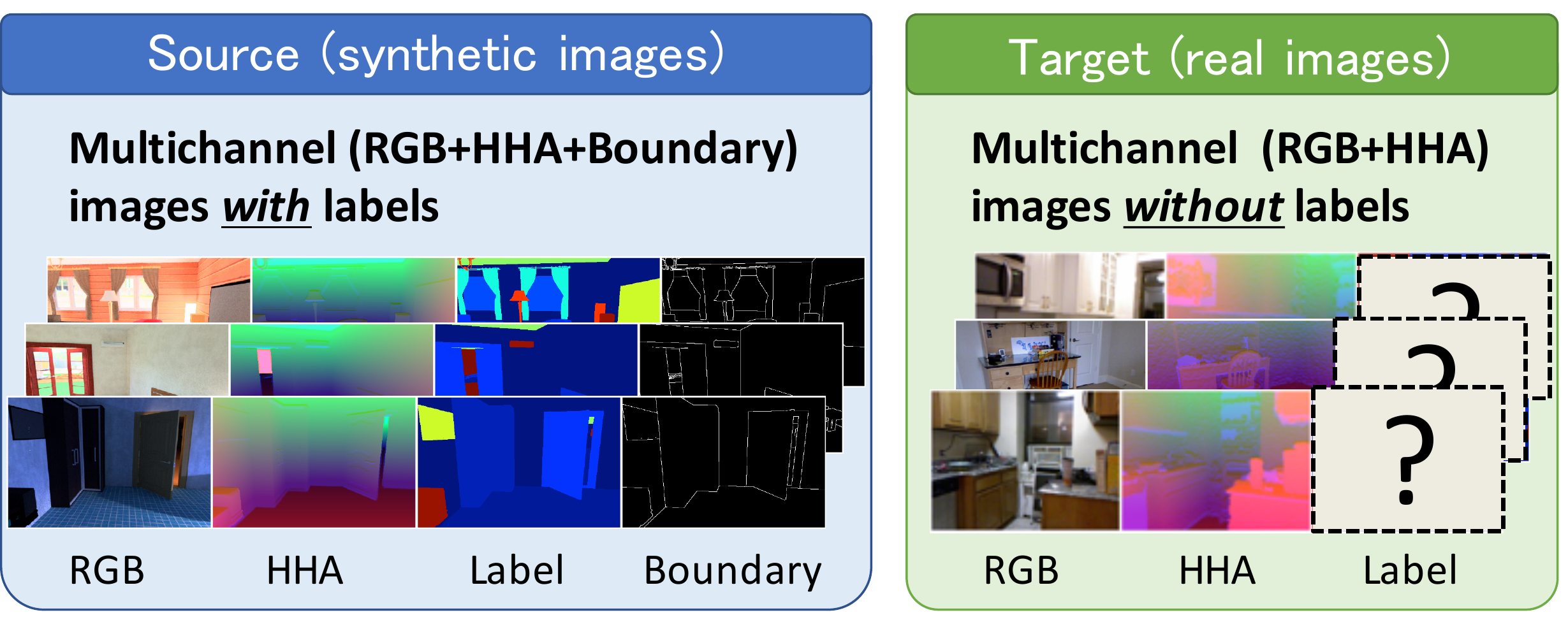}
\caption{Setting of this research. Left is samples of SUNCG~\cite{zhang2017physically} and right is samples of NYUDv2~\cite{silberman2012nyu}. HHA is the three dimensional encoding of depth~\cite{gupta2014learning}.}
\label{fig:setting}
\end{figure}

We propose two approaches that can efficiently use multichannel inputs with an UDA algorithm.
One is a fusion-based approach that uses different modal images as inputs and the other is a multitask learning approach that uses only RGB images as inputs but other modal images as outputs.
Fusing different modalities (RGB, depth or boundary) efficiently is known to boost the segmentation accuracy compared to a simple concatenation of inputs known as \textit{early fusion} in past research.
Except for \textit{early fusion}, many fusion methods~\cite{FuseNet,cheng2017locality,park2017rdfnet,haIROS2017} exist and their efficacy is task-specific, which makes us rethink their ideas.
Multitask learning is also a promising approach. Multitask learning that solves related tasks such as semantic segmentation and depth estimation tasks simultaneously is known to boost each task's performance~\cite{kendall2017multi,kuga2017multi}.
In multitask learning it is easy to add another single task, such as a boundary detection task, which can be thought to render feature maps more aware of boundaries.
Boundary detection output can be utilized collaterally to refine the messy domain-adapted segmentation output.

In summary, the specific contribution of this paper includes:
\begin{itemize}
\item We combine a multichannel semantic segmentation task with an unsupervised domain adaptation (UDA, see \figref{fig:setting}) task and propose two approaches (fusion-based and multitask learning)
 \item We show that the multitask learning approach outperforms the simple \textit{early fusion} approach according to all evaluation metrics.
 \item We propose adding a boundary detection task to the multitask learning approach and use the detection result to refine the segmentation output, which improves both the qualitative and quantitative results.
\end{itemize}


\section{Related work}
Here, we describe two related research themes, \textit{domain adaptation for semantic segmentation} and \textit{semantic segmentation with multichannel image}.

\begin{figure}[t]
  \centering
  \begin{minipage}[b]{.5\linewidth}
    \includegraphics[width=0.8\hsize]{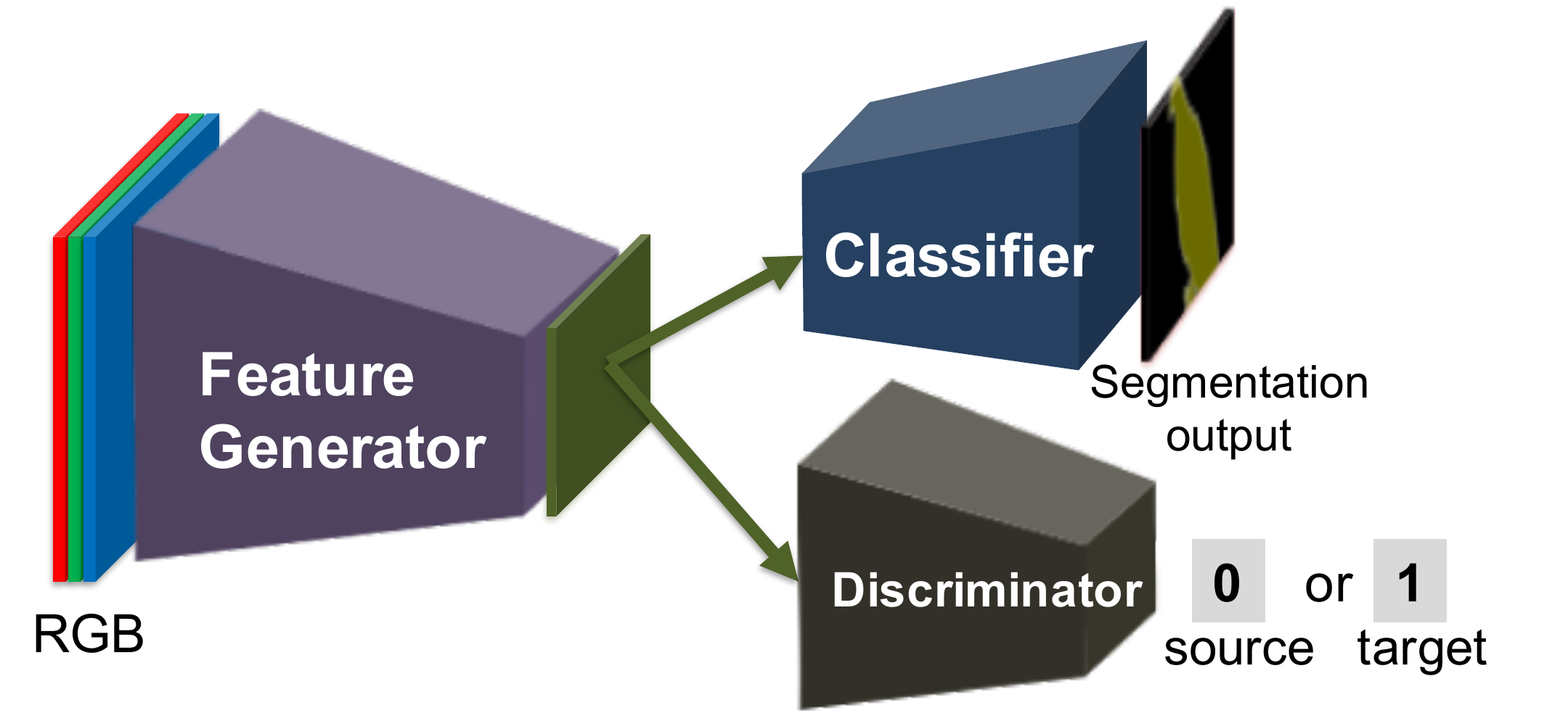}
    \subcaption{DANN}
    \label{fig:dann}
  \end{minipage}%
  \begin{minipage}[b]{.5\linewidth}
    \includegraphics[width=.8\hsize]{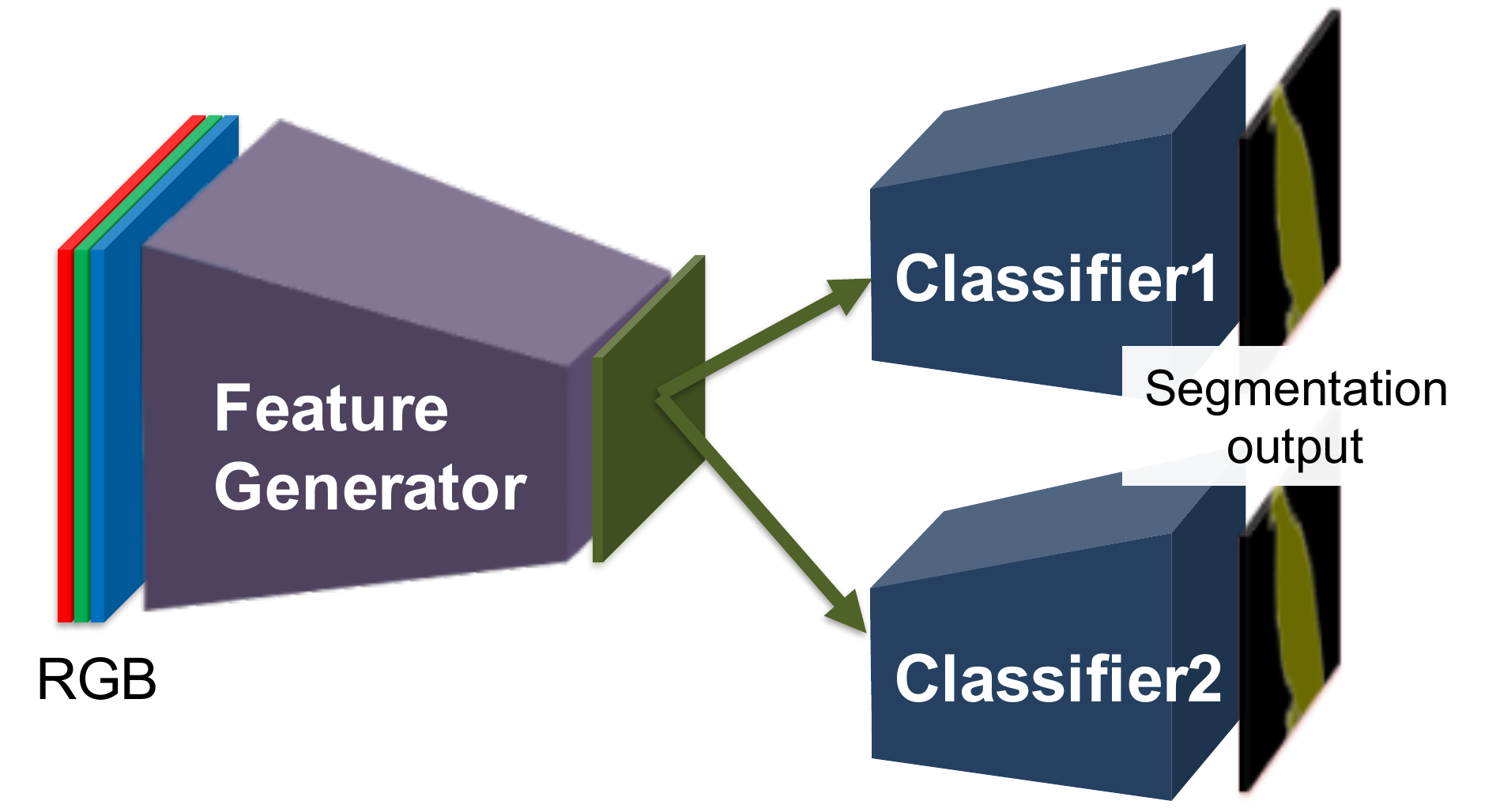}
    \subcaption{MCD}
    \label{fig:mcd}
  \end{minipage}
  \caption{Example of network architectures for domain adaptation [(a) domain adversarial neural network (DANN)~\cite{hoffman2016fcns} and (b) maximum classifier discrepancy (MCD)~\cite{saito2017maximum}].
  When inputs belong to source samples which have labels, we train a feature generator and a classifier using the labels.
  When inputs belong to target samples which have \textbf{no} labels, we train a feature generator and (one discriminator in the case DANN is used or two classifiers in the case MCD is used) in an adversarial manner.}
  \label{fig:adversarial_architectures}
\end{figure}

\subsection{Domain adaptation for semantic segmentation}
\label{sec:da_seg}

When we train a classifier in one (source) domain and apply it to classify samples in a new (target) domain, the classifier is known not to generalize well in the new domain due to the domain's difference. Many methods tackle the problem by aligning distributions of features between the source and target  domain~\cite{ganin2014unsupervised,long2015learning,tzeng2017adversarial,bousmalis2016unsupervised}.
These methods are proposed to deal with classification problem. Recently, methods for semantic segmentation have been proposed too.
Hoffman \etal\cite{hoffman2016fcns} first tackled this problem. They adopted an adversarial training framework which has a feature extractor and a discriminator (see \figref{fig:dann}). A discriminator tries to detect whether the extracted feature correctly comes from source samples or target samples, while a feature extractor tries to generate features that deceive a discriminator in an adversarial manner. The other researches on this theme also leverage adversarial training.
Zhang \etal\cite{Zhang_2017_ICCV} adopted curriculum learning that starts the easier task (global and super pixel label distribution of source samples matching those of target samples), then tries to solve difficult tasks (semantic segmentation).
Cheng \etal\cite{chen2017no} tried to tackle the cross-city adaptation problem via adversarial training and extract static-object priors that can be obtained from the Google Street View time-machine feature.

We utilize Saito \etal\cite{saito2017maximum}'s method (MCD), which is shown to be effective in  segmentation task. They proposed a method that uses two classifiers' difference of output (called discrepancy) to align features between the source and target domain. They trained one feature extractor network and two different classifier networks for the same task (see \figref{fig:mcd}). Two classifiers are trained to increase the discrepancy for target samples whereas feature extractor is trained to decrease it.  Details are in \secref{sec:annotation_free}.

\subsection{Semantic segmentation with multichannel (RGBD) image}

Previously, RGBD segmentation was conducted based on handcrafted features specifically designed for capturing depth as well as color features~\cite{gupta2013perceptual}.
Long \etal\cite{FCN} proposed a fully convolutional neural network (FCN) for semantic segmentation. FCN not only replaced the fully connected layer in classification models such as AlexNet~\cite{AlexNEt} with a convolutional layer but also proposed using two methods, \textit{deconvolution} and \textit{shortcut}, which are now widely used in many semantic segmentation models. Long \etal also reported the segmentation scores of the NYUDv2 dataset~\cite{silberman2012nyu}, where RGB + Depth were combined in the input (we call this \textit{early fusion}) and RGB + HHA in output (we call this \textit{score fusion}). HHA is the three dimensional encoding of depth (Horizontal disparity, Height above ground, and the Angle of the local surface normal with the inferred gravity direction) proposed by Gupta \etal\cite{gupta2014learning}.
FuseNet~\cite{FuseNet} prepared an RGB and depth encoder separately then fused the two encoders in certain middle layers (see \figref{fig:fusenet_like_fusion}).
The locality sensitive deconvolutional network with Gate Fusion~\cite{cheng2017locality} used an affinity matrix embedded with pairwise relations between neighboring RGB-D pixels to recover sharp boundaries of FCN maps. Gate Fusion (see \figref{fig:score_fusion}) learns to adjust to the contributions of RGB and depth that exist in the last layer of the network.
RDFNet~\cite{park2017rdfnet} fuses two networks with multi-modal feature fusion blocks and multi-level feature refinement blocks following RefineNet~\cite{lin2016refinenet}.

The above-mentioned approaches utilize all the different modals as input, but there is also an approach that utilizes only RGB as input and the other modals as output; this is the multitask learning approach.
Multitask learning is a promising approach for efficiently and effectively addressing multiple mutually-related recognition tasks and its performance is known to outperform that of the single task methods. Kendall \etal worked on three tasks (semantic and instance segmentation, and depth estimation)~\cite{kendall2017multi} and Kuga \etal also worked on three tasks (RGB reconstruction, semantic segmentation, and depth estimation)~\cite{kuga2017multi}.

There are other approaches using geometric cues obtained from depth images~\cite{qi20173d,lin2017cascaded}, but in this research, we just focus on fusion-based and multitask learning approaches, which renders our model not only applicable to geometric applications but also other modal images, such as thermal images.

\section{Proposed models}
Our objective is to conduct unsupervised semantic segmentation with multichannel input.
In order to realize that, the two required functionalities are:

\begin{itemize}
\item Annotation free (using no labels in a target dataset)
\item Using different modalities [RGB, Depth (HHA), (Boundary)] efficiently
\end{itemize}

\subsection{Annotation free (using no labels in a target dataset)}
\label{sec:annotation_free}
\begin{figure}[t]
  \centering
\includegraphics[width=0.8\hsize]{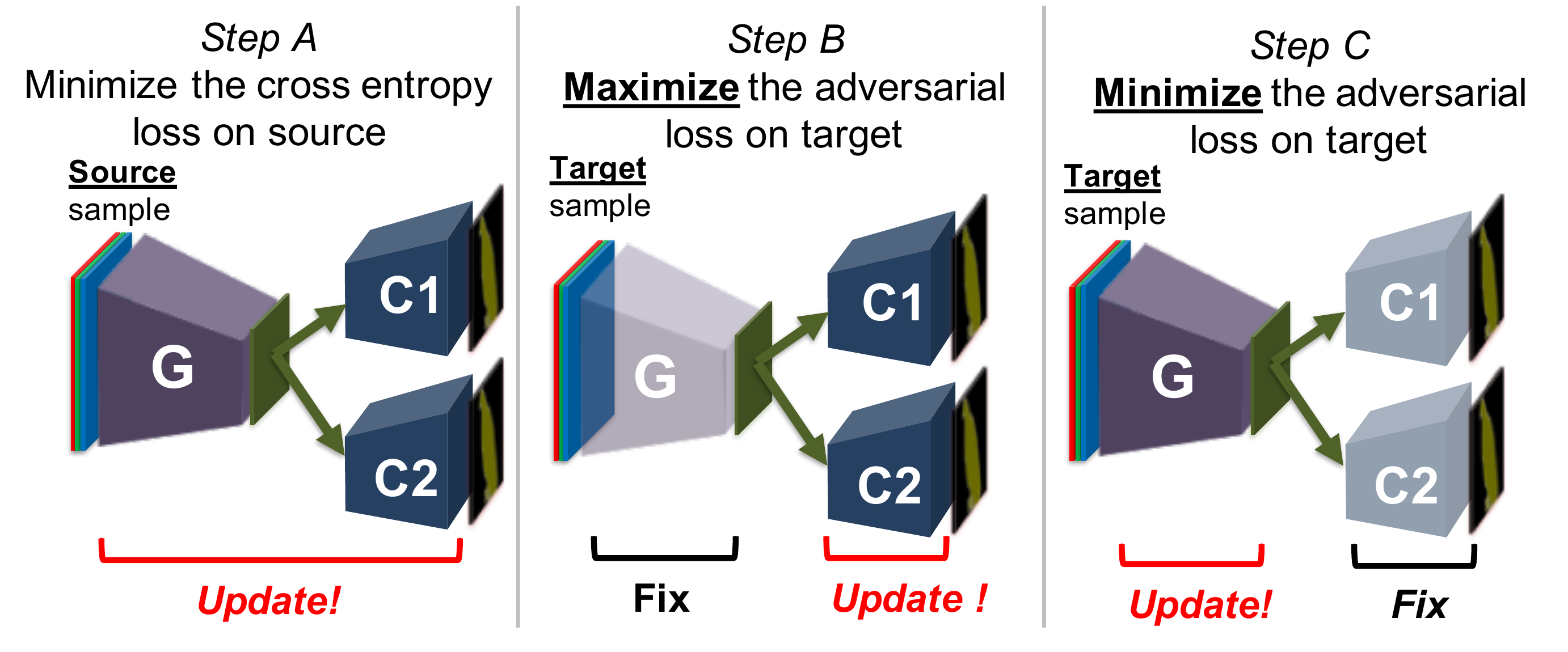}
\caption{Adversarial training steps in MCD~\cite{saito2017maximum}.}
\label{fig:mcd_algorithm}
\end{figure}

In order to satisfy the former function, a simple solution utilizes an existing dataset or synthetic dataset which we can generate on our own.
However, if there is a domain shift (a difference of appearance or label distribution) between existing training data and test data, the performance can be poor.
We tackle this case because a domain shift usually exists between synthetic and real datasets.
Hence, we use a domain adaptation algorithm, which leverages adversarial training and enables the model to extract domain-robust features.
In order to adopt the adversarial training algorithms, we separate an end-to-end segmentation model into a feature generator and a classifier as shown in \figref{fig:adversarial_architectures}.
To utilize MCD~\cite{saito2017maximum}, we prepare two classifiers ($C_1$, $C_2$) and train them by three steps as shown in \figref{fig:mcd_algorithm}.

\newcommand{\mymin}{\mathop{\rm min}\limits}
\newcommand{\mymax}{\mathop{\rm max}\limits}
\newcommand{\1}{\mbox{1}\hspace{-0.25em}\mbox{l}}
\newcommand{\bx}{\textbf{x}}
\newcommand{\by}{\textbf{y}}
\newcommand{\bX}{\textbf{Y}}
\newcommand{\bY}{\textbf{Y}}

\noindent \textbf{Formulation:}
We have access to a labeled source RGB image $\bx^{s}_{\rm RGB}$, HHA image $\bx^{s}_{\rm HHA} (=\by^{s}_{2})$, instance boundary image $\by^{s}_{3}$ and a corresponding semantic segmentation label $\by^{s}_{1}$ drawn from a set of labeled source images \{$X^{s}_{\rm RGB}$, $Y^{s}_{1}$,
$Y^{s}_{2}(= X^{s}_{\rm HHA})$, $Y^{s}_{3}$\},
as well as an unlabeled target image $\bx^{t}_{\rm RGB},\bx^{t}_{\rm HHA}$ drawn from unlabeled target images \{$X^{t}_{\rm RGB}, Y^{t}_{2}(= X^{t}_{\rm HHA})$\}.
We train a feature generator network $G$, which takes inputs $\bx^{s}$ or $\bx^{t}$, and classifier networks $C_1$ and $C_2$, which take features from $G$. 
$C_1$ and $C_2$ classify them into $K$ classes per pixel, that is, they output a $(K\times|\bx|)$-dimensional vector of logits. Note that $|\bx|$ denotes the number of pixels per image. We obtain class probabilities by applying the softmax function for the vector. We use the notation $p_{1}(\by_{1}|\bx)$, $p_{2}(\by_{1}|\bx)$ to denote the $(K\times|\bx|)$-dimensional probabilistic outputs for input $\bx$ obtained by $C_1$ and $C_2$ respectively.

\subsubsection{Step A}
We train $G$, $C_1$ and $C_2$ to classify the source samples correctly. In order to make classifiers and generator obtain task-specific discriminative features, this step is crucial. We train the networks to minimize softmax cross entropy. The objective is as follows:

\footnotesize
\begin{equation}
   \mymin_{G,C_1,C_2} \mathcal{L}_{\rm seg}(X^{s}_{\rm RGB},Y^{s}_{1})
\end{equation}
\begin{equation}
 \mathcal{L}_{\rm seg}(X^{s}_{\rm RGB},Y^{s}_{1}) = Y^{s}_{1} \log p(Y^{s}_{1} | X^{s}_{\rm RGB})
 \label{eq:crossentropy}
\end{equation}
\normalsize

\subsubsection{Step B}
We train $C_1$ and $C_2$ as a discriminator with fixing $G$.
Let $\mathcal{L}_{\rm adv}(X_{t})$ be the adversarial loss that can be computed using target sample. This loss measures the discrepancy of $C_1$ and $C_2$.
A classification loss on the source samples is also added for better performance.
The same number of source and target samples were randomly chosen to update the model.
The objective is as follows:
\footnotesize
\begin{equation}
  \mymin_{C_1,C_2} \mathcal{L}_{\rm seg}(X^{s}_{\rm RGB},Y^{s}_{1}) - \mathcal{L}_{\rm adv}(X^{t}_{\rm RGB}). \\
\end{equation}
\normalsize

\subsubsection{Step C}
We train $G$ to minimize the adversarial loss with fixing $C_1$ and $C_2$.
The objective is as follows:
\footnotesize
\begin{equation}
 \mymin_{G} \mathcal{L}_{\rm adv}(X^{t}_{\rm RGB}).\\
\end{equation}
\normalsize
The target and source images feed to the training randomly and these three steps are repeated until convergence of all the parts (classifiers and generator). The order of the three steps is not important but it is important to train the classifiers and generator in an adversarial manner under the condition that they can classify source samples correctly.

However, this still outputs messy segmentation results for the indoor scene recognition task (see \figref{fig:sample_results}).
We propose to refine this by using boundary detection output that can be gained via a multitask learning approach (Details are in \secref{sec:multitask})

\subsection{Using different modalities [RGB, Depth (HHA), (Boundary)] efficiently}
In order to satisfy the latter function, we propose the two approaches below:
\begin{enumerate}
\item Fusion-based approach that uses all different modal images as input
\item Multitask learning approach that uses only RGB as input and the other modals as output.
\end{enumerate}

We will describe these two approaches in detail.

\begin{figure}[t]
  \centering
\begin{minipage}[b]{.3\linewidth}
  \centering
    \includegraphics[width=.9\hsize]{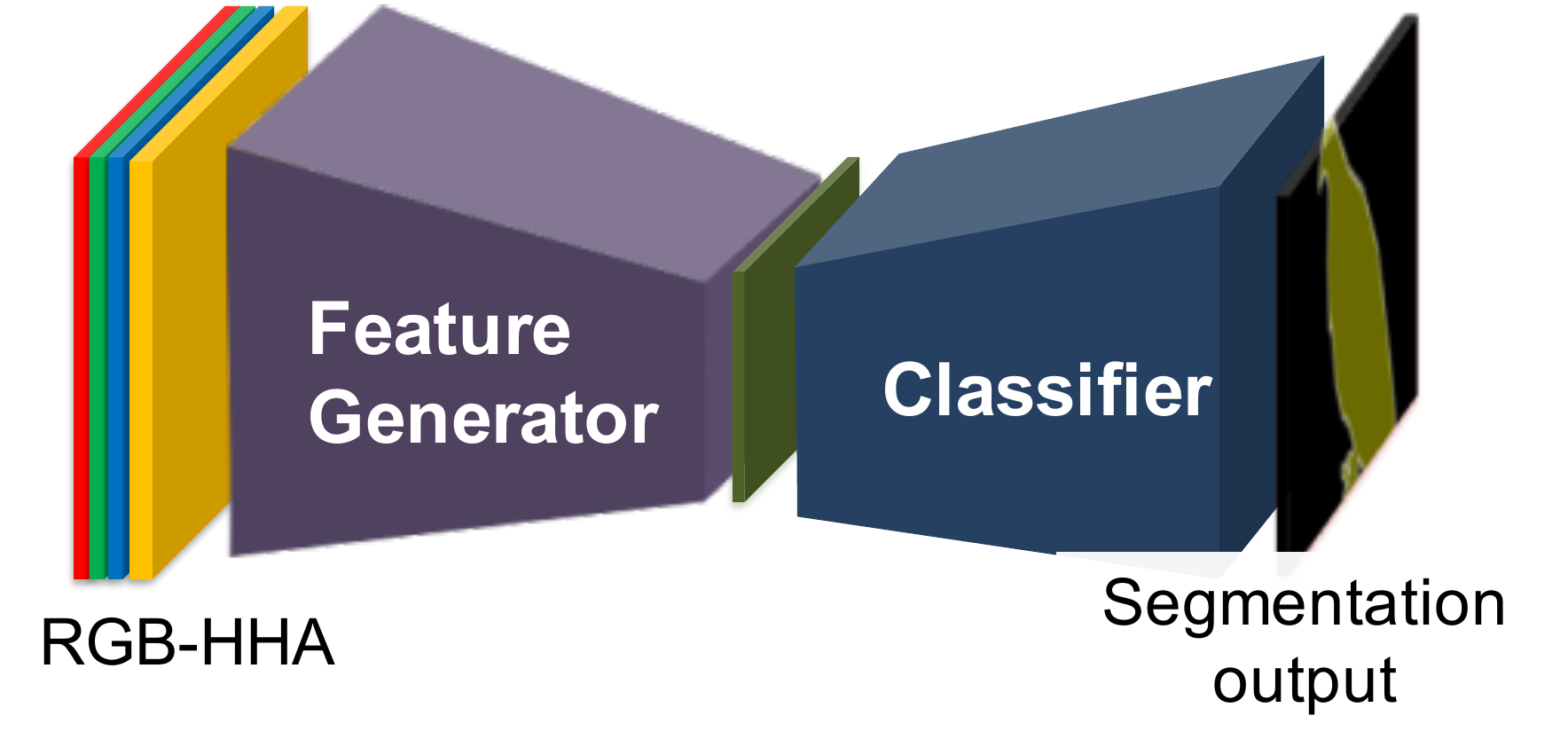}
    \subcaption{Early Fusion}
    \label{fig:early_fusion}
\end{minipage}
\begin{minipage}[b]{.3\linewidth}
  \centering
  \includegraphics[width=.9\hsize]{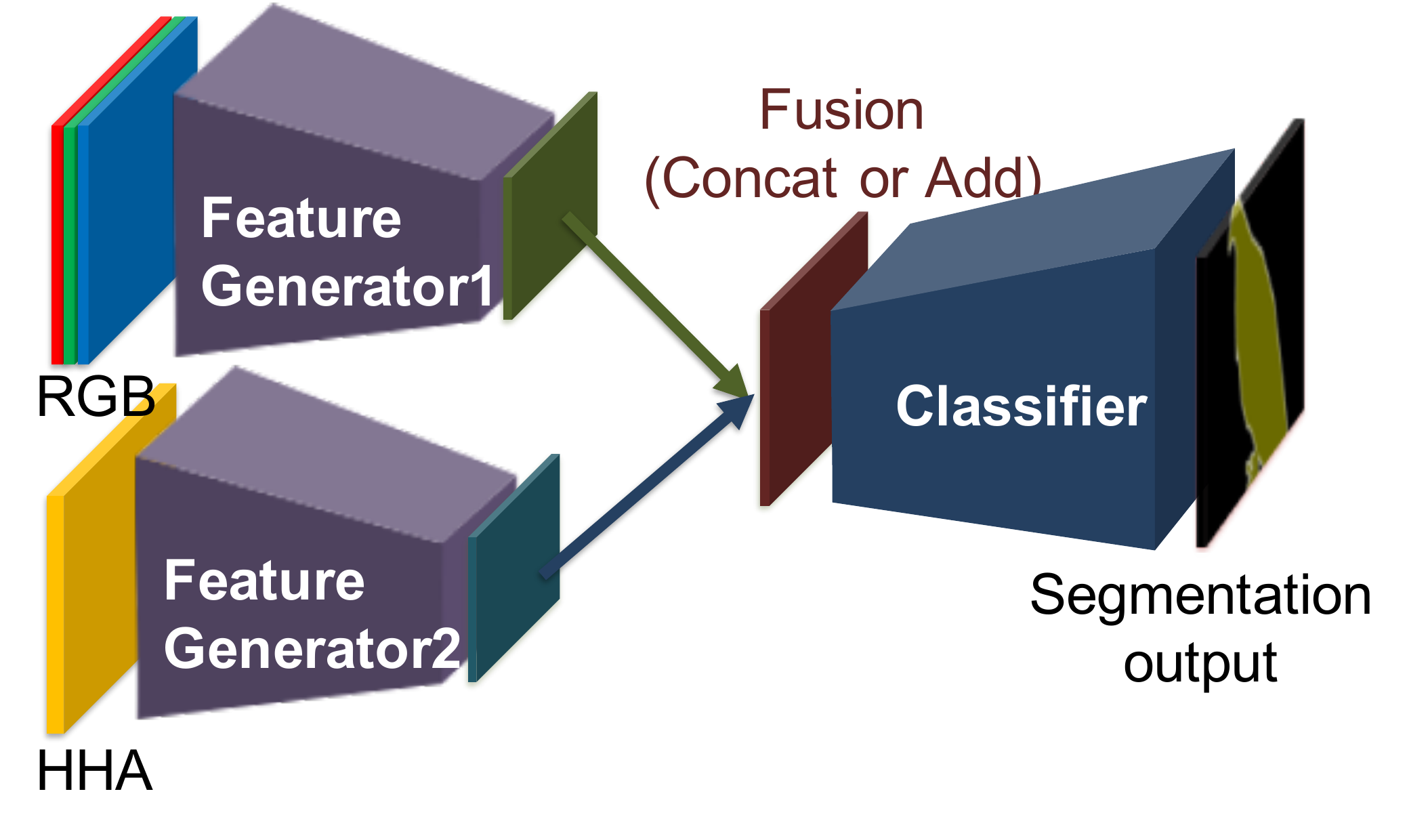}
  \subcaption{Late Fusion}
  \label{fig:late_fusion}
\end{minipage}
\begin{minipage}[b]{.3\linewidth}
  \centering
  \includegraphics[width=.9\hsize]{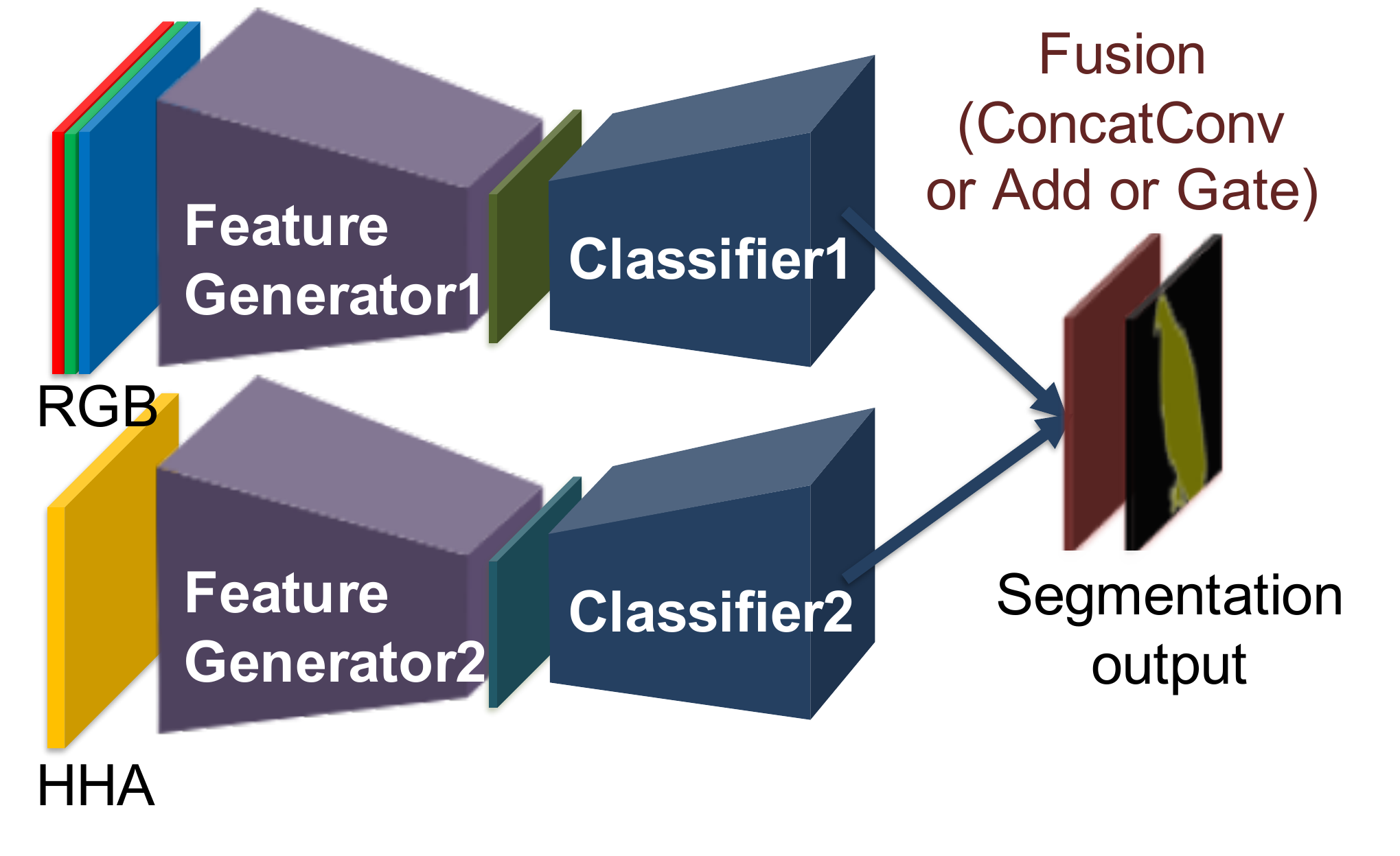}
  \subcaption{Score Fusion}
  \label{fig:score_fusion}
\end{minipage} \\
\begin{minipage}[b]{.3\linewidth}
  \centering
  \includegraphics[height=0.6\linewidth]{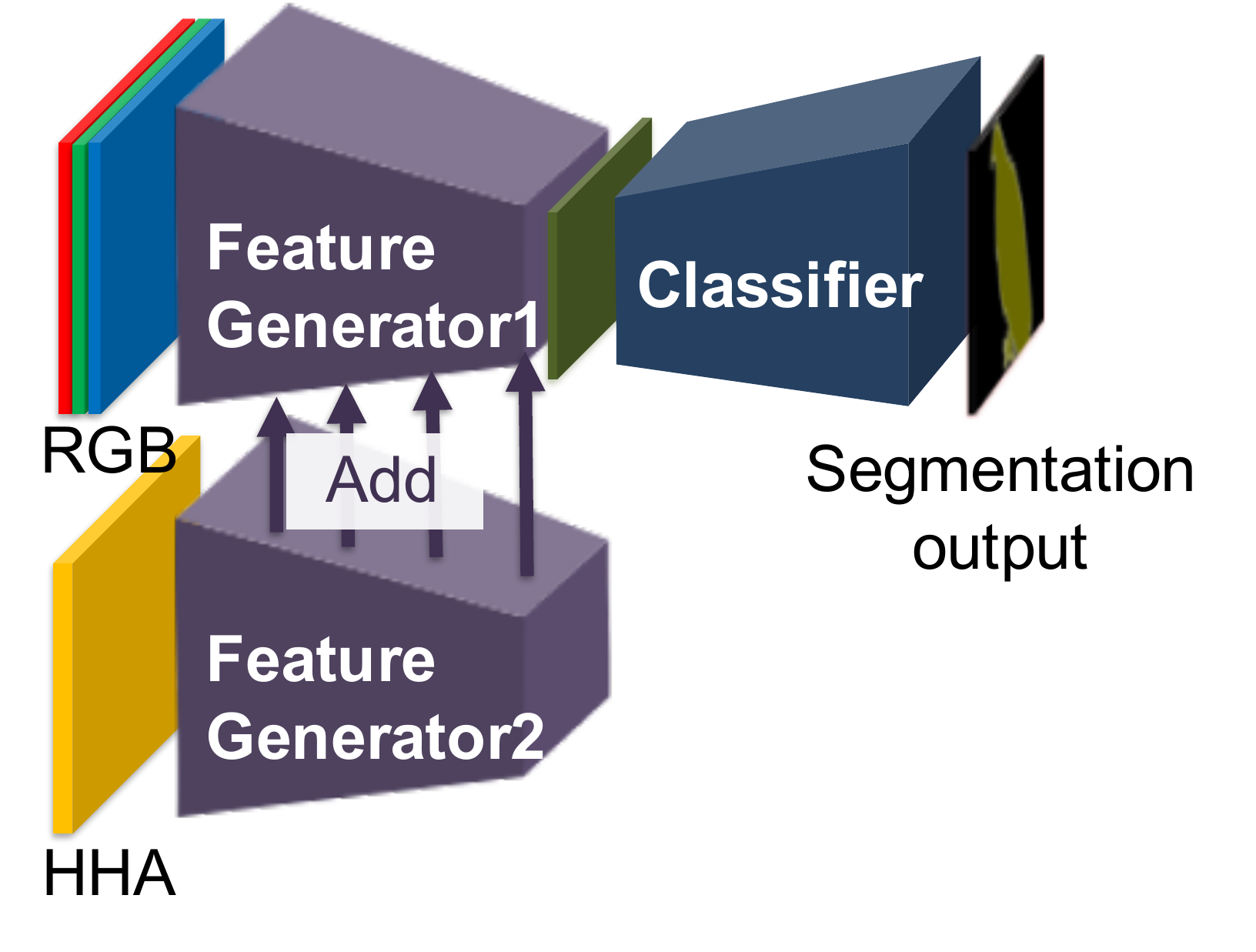}
  \subcaption{FuseNet like Fusion}
  \label{fig:fusenet_like_fusion}
\end{minipage}
\begin{minipage}[b]{.3\linewidth}
  \includegraphics[width=.7\hsize]{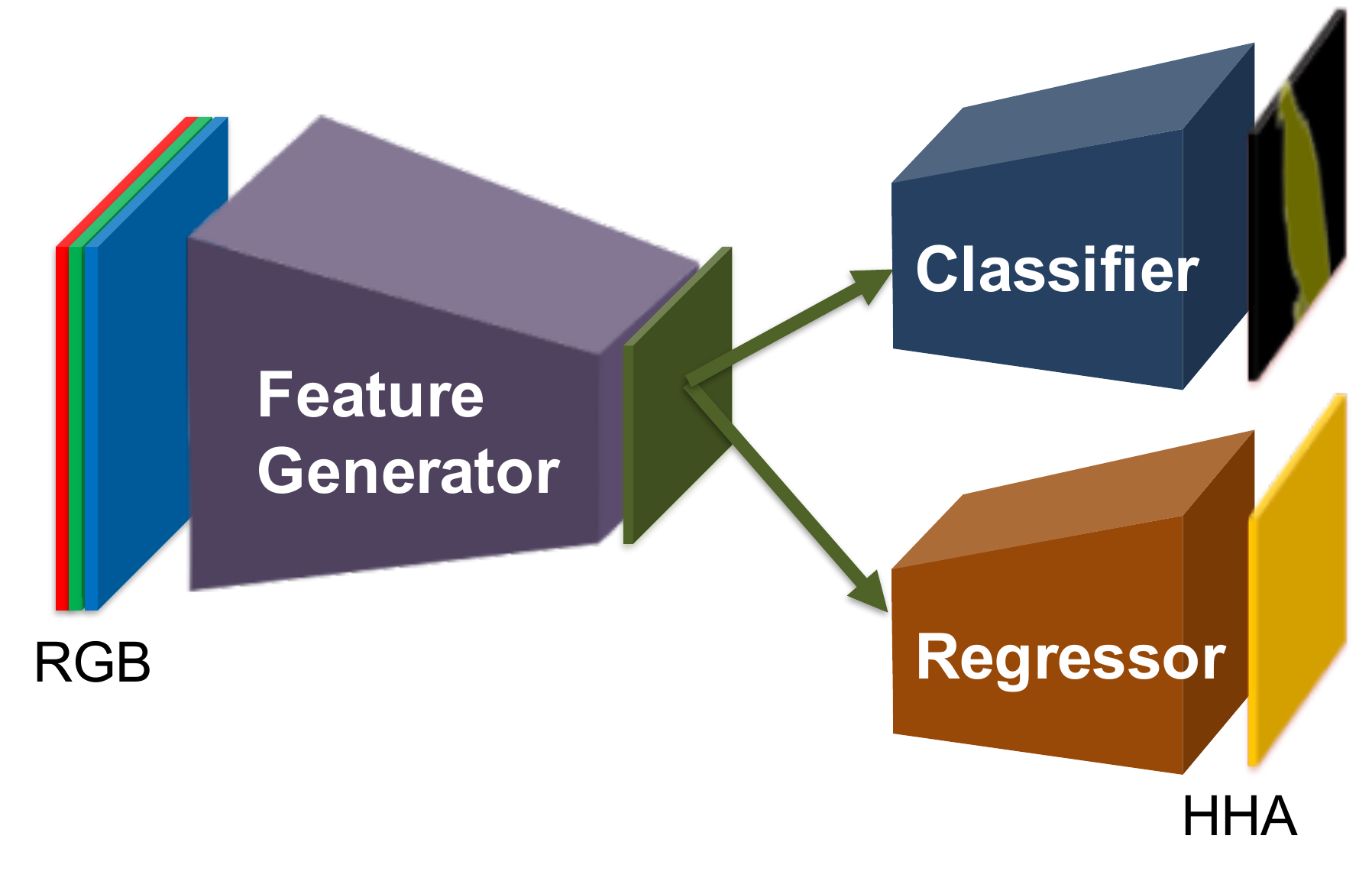}
  \vspace{10pt}
  \subcaption{Multitask (Dual)}
  \label{fig:dual_task}
\end{minipage}%
\begin{minipage}[b]{.3\linewidth}
  \includegraphics[width=.7\hsize]{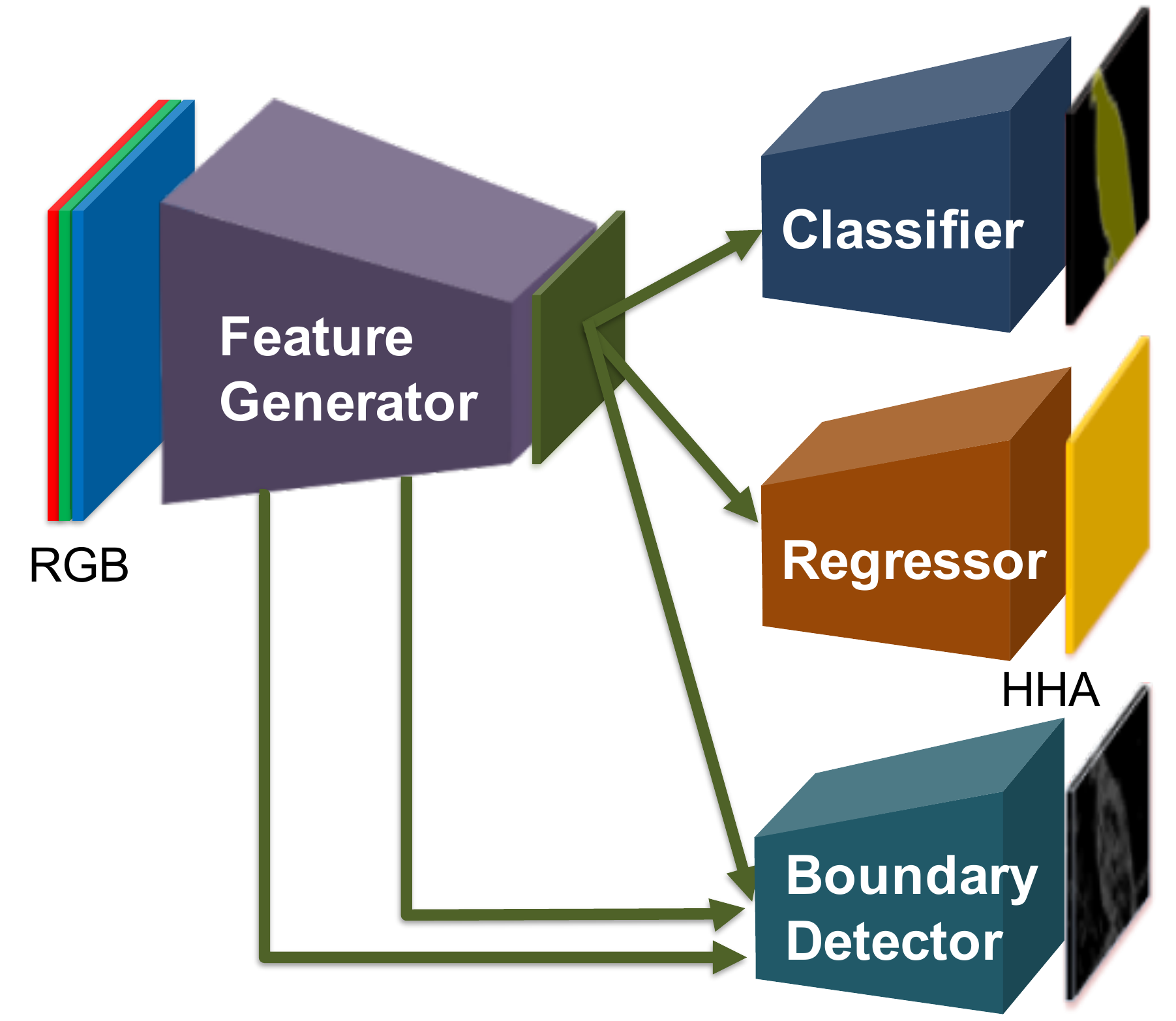}
  \subcaption{Multitask (Triple)}
  \label{fig:triple_task}
\end{minipage} \\

\caption{Fusion-based [(a)-(d)] and multitask learning [(e), (f)] architectures. (e) model that solves the semantic segmentation and depth regression task and (f) model that solves the boundary detection task in addition to the other two tasks. In this figure, one classifier (two classifiers in the score fusion model) exists but actually two classifiers (four classifiers in the score fusion model) exist to utilize MCD as shown in \figref{fig:mcd}.}
\label{fig:architectures}
\end{figure}


\subsubsection{Fusion-based approach}

If multimodal images are inputs, an appropriate fusing method is known to boost segmentation accuracy.
There are many fusion methods~\cite{FuseNet,cheng2017locality,park2017rdfnet,haIROS2017} but in this research we focus on four comparatively simple fusions that are \textit{early fusion}, \textit{late fusion}, \textit{score fusion}, \textit{fusenet like fusion}(see \figref{fig:early_fusion}~-~\figref{fig:fusenet_like_fusion}), because they are not specifically designed and widely used.
\textit{Early fusion} just concatenates the RGB and HHA (depth) in inputs.
\textit{Late fusion} fuses two encoders in the middle (in this research, the middle means the one layer before the final output). When fusing, we consider two ways of fusing, addition or concatenation.
\textit{Score fusion} fuses two encoders in the output. When fusing, we also consider three fusing methods, addition or concatenation + $1 \times 1$ convolution or gate fusion~\cite{cheng2017locality}.
\textit{Fusenet-like fusion} fuses two encoders in certain middle layers~\cite{FuseNet}.
Past research~\cite{hoffman2016cross,song2017depth} showed that lower layers of a CNN are largely task and category agnostic but domain-specific, while higher layers are largely task and category specific but domain-agnostic.
So \textit{late fusion} is considered to be the best out of the four.

To incorporate these fusion-based models into MCD algorithm which we described in \secref{sec:annotation_free}, we just replace $X_{\rm RGB}$ with $X_{\rm RGBHHA}$.

\subsubsection{Multitask learning approach}
\label{sec:multitask}

Multitask learning is a promising approach for efficiently and effectively addressing multiple mutually related recognition tasks and its performance is known to outperform that of single tasks~\cite{kendall2017multi}.
We solve semantic segmentation and depth regression tasks simultaneously.
Also, a general segmentation model is known to get not sharp boundaries~\cite{cheng2017locality,park2017rdfnet}.
Kendall \etal\cite{kendall2017uncertainties} showed that the points on object boundaries have high aleatoric uncertainty.
Hence, we add one extra task, a boundary detection task, which can be thought to render feature maps more aware of boundaries.
One feature map is used as input for the semantic segmentation and depth estimation task. In addition, two lower feature maps were also used as inputs for the boundary detector following holistically-nested edge detection (HED)~\cite{xie2015holistically} as shown in \figref{fig:triple_task}.

\begin{figure}[t]
  \centering
\includegraphics[width=0.7\hsize]{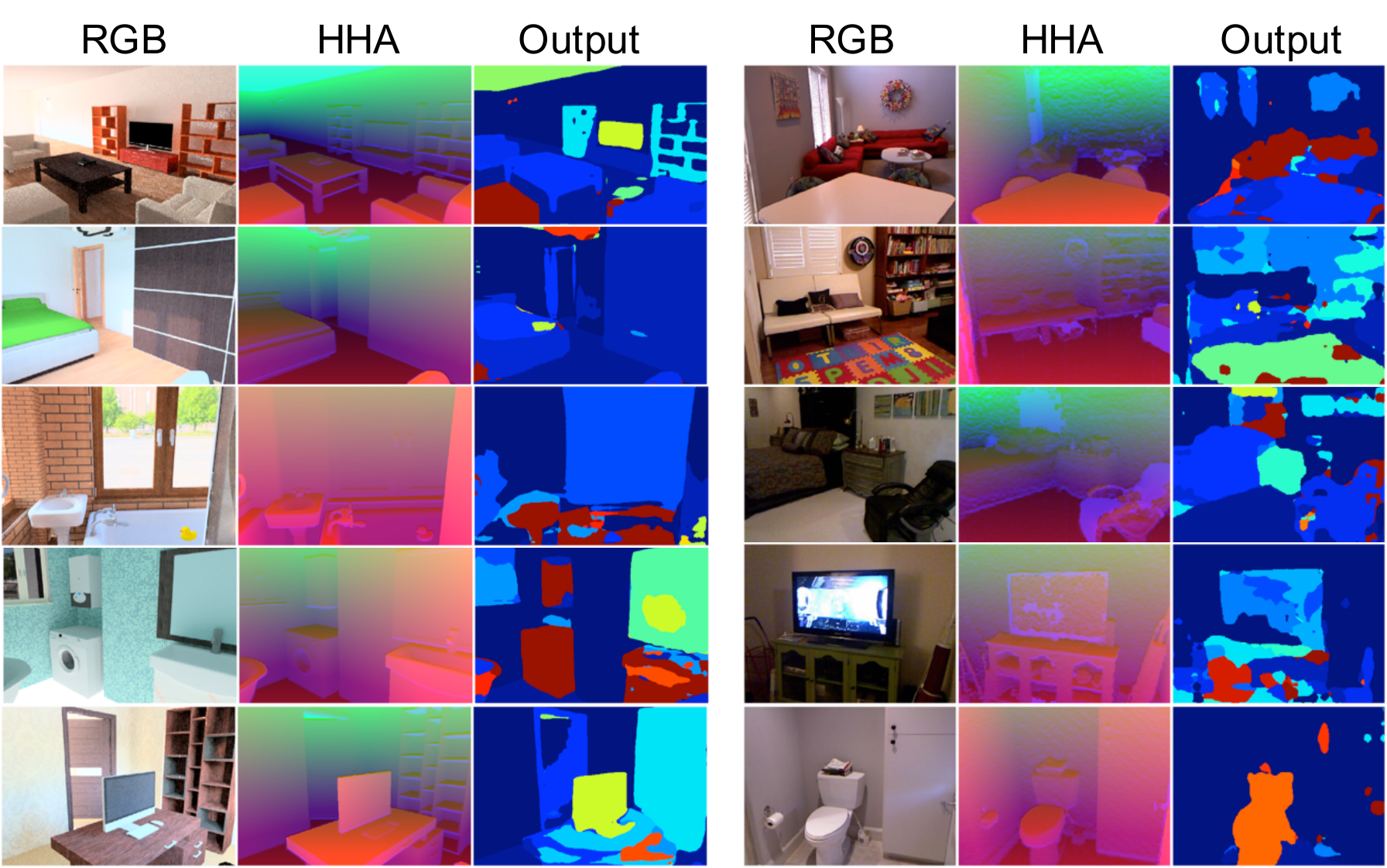}
\caption{Sample segmentation results. \textbf{Left:} Output of the model that was trained on SUNCG and tested on SUNCG. \textbf{Right:} Output of the domain-adapted model from SUNCG to the NYUDv2 train split, and tested on the NYUDv2 test split. The domain adapted model outputs messier results than usual.}
\label{fig:sample_results}
\end{figure}

\begin{figure}[t]
  \centering
\includegraphics[width=0.7\hsize]{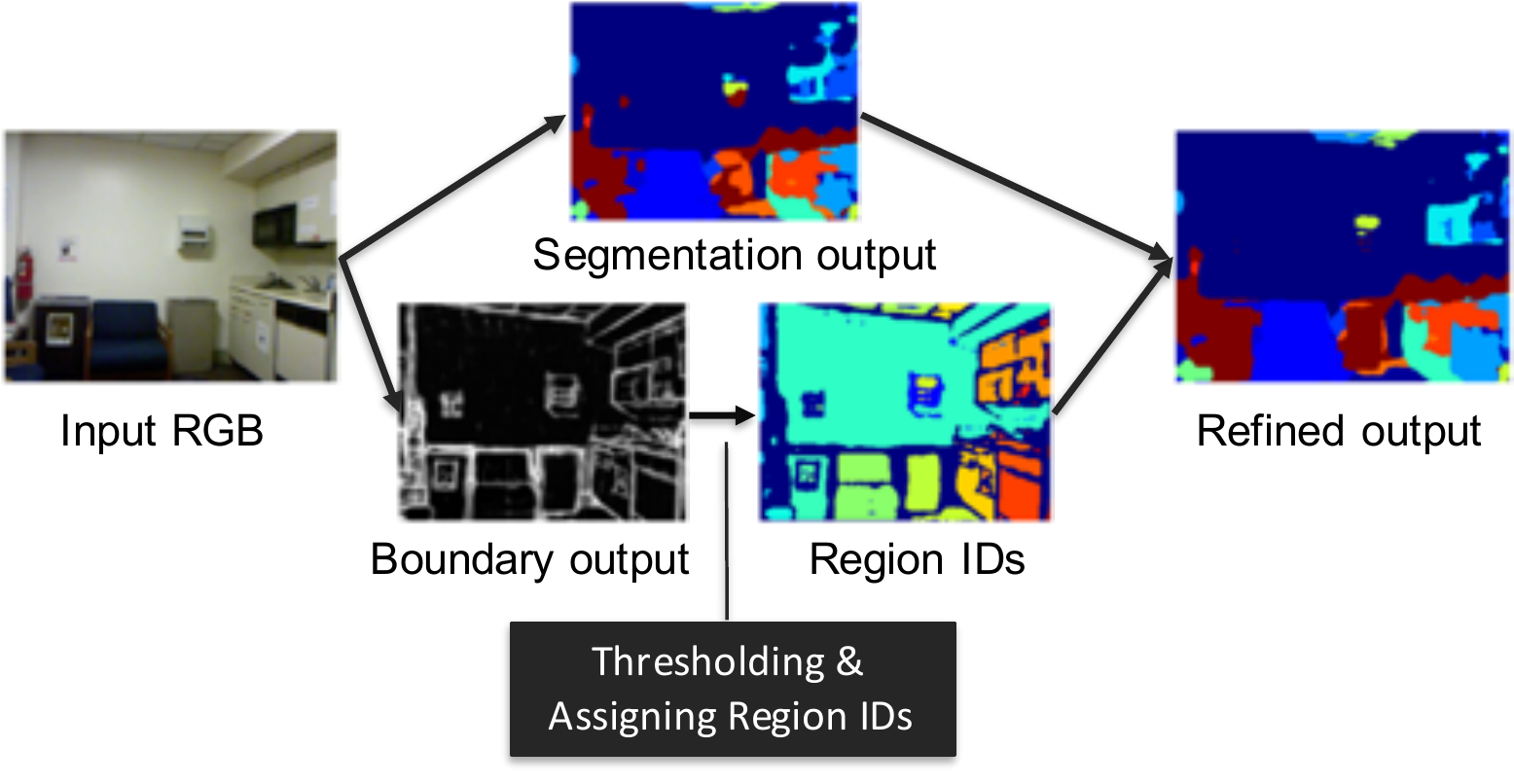}
\caption{Refinement of segmentation result using boundary detection output.}
\label{fig:postprocess}
\end{figure}

This boundary detection output can be utilized to refine the segmentation output.
In fact, based on the segmentation results in \figref{fig:sample_results}, the outputs of the domain-adapted model are messier than usual.
In order to fix this, we propose post-processing the segmentation result based on boundary detection output as shown in \figref{fig:postprocess}.
In detail, we first threshold the boundary detection output and then assign IDs to each separated region. Segmentation output (class label) in one region should be unique (one class label for one ID) and so the segmentation output is refined by voting in each separated region.
However, boundary detection output is not always perfect. One region sometimes expands to the adjacent region and becomes too large.
Therefore, we do not post-process a region whose area is bigger than the maximum-threshold (set to one-third of the image size). In addition, points exactly on the boundaries are not post-processed.

To incorporate these multitask learning models into MCD algorithm which we described in \secref{sec:annotation_free}, we replace the semantic segmentation loss of \textit{Step A} with a total multitask loss.
When we compute the total loss, tuning the weight of each task is important.
We adopted Kendall's algorithm to automatically tune the weight~\cite{kendall2017multi} by introducing trainable homoscedastic uncertainty parameter $\sigma_{i}$, where $i$ denotes the task index (1: semantic segmentation, 2: depth regression, 3: boundary detection).
This total loss is computed as follows;

\footnotesize
\begin{align}
  \begin{array}{l}
  \mathcal{L}_{\rm multitask}(X^{s}_{\rm RGB},X^{t}_{\rm RGB},Y^{s}_{1},Y^{s}_{2},Y^{t}_{2} (, Y^{s}_{3}))  \notag \\
  = \sum_{i \in \{1,2(,3)\}} \left(\frac{1}{2 \sigma_i^2} \mathcal{L}^{s}_{i}(X^{s}_{\rm RGB},Y^{s}_{i}) + \log \sigma_{i}^{2} \right)
  + \mathcal{L}^{t}_{2}(X^{t}_{\rm RGB},Y^{t}_{2})
  \end{array}
\end{align}
\normalsize

\noindent where $\mathcal{L}_{1}(=\mathcal{L}_{seg})$, $\mathcal{L}_{2}$, $\mathcal{L}_{3}$ denotes the cross entropy loss for semantic segmentation, the mean squared loss for depth (HHA) regression loss, the class-balanced cross entropy loss for boundary detection, respectively. When $\mathcal{L}_{3}$ is not used, the model corresponds to \figref{fig:dual_task} and, otherwise it corresponds to  \figref{fig:triple_task}. Note that only depth regression loss ($\mathcal{L}_{2}$) is computed on both source and target samples but segmentation and boundary detection losses are computed only on source samples because we hypothesize that semantic labels and instance boundaries only exist in source samples.
$\mathcal{L}_{1}(=\mathcal{L}_{seg})$ is computed as \equref{eq:crossentropy} and $\mathcal{L}_{2}$, $\mathcal{L}_{3}$ are computed as follows;

\footnotesize
\begin{align}
  \mathcal{L}_{2} =&~||Y^2_{2} - f(X_{\rm RGB}) ||^2 \\
  \mathcal{L}^{s}_{3} =& - \frac{|Y^{s}_{3-}|}{|Y^{s}_{3}|} \sum_{j \in Y^{s}_{3+}} \log p(y^{s}_{3j}=1| X_{\rm RGB})
                        - \frac{|Y^{s}_{3+}|}{|Y^{s}_{3}|} \sum_{j \in Y^{s}_{3-}} \log p(y^{s}_{3j}=0| X_{\rm RGB})
\end{align}
\normalsize

\noindent where $f$ transforms input $X_{\rm RGB}$ to depth (HHA) regression output, and $|Y_{3-}|$ and $|Y_{3+}|$ denote the edge and non-edge ground truth label sets, respectively. (Note that $Y^{s}_{3} \in \{0,1\}$ and $|Y_{3}| = |Y_{3+}| + |Y_{3-}|$.) $\log p(y^{s}_{3j}=1| X_{3})$ and $\log p(y^{s}_{3j}=0| X_{3})$ denotes the sigmoid output of predicted boundary detection on the edge and non-edge points, respectively.

\section{Experiment}
\subsection{Setting}
\noindent \textbf{Implementation detail\footnote{our code: https://github.com/LittleWat/multichannel-semseg-with-uda}:}
We use a dilated residual network (\textit{drn\_d\_38})~\cite{Yu2017} which is pre-trained on ImageNet~\cite{deng2009imagenet}, which was shown to perform well in \cite{saito2017maximum}.
We followed the public implementation\footnote{https://github.com/fyu/drn} and adopted MCD~\cite{saito2017maximum} as an unsupervised domain adaptation method because it had good performance on domain adaptation problems from synthetic GTA~\cite{richter2016playing} to real CityScapes~\cite{CityScapes}. Then, we separated \textit{drn\_d\_38} into a feature generator and a classifier (actually two classifiers) and trained them in an adversarial manner.
In fusion-based models, the last transposed convolution layer was used as a classifier and all lower layers were used as feature generators.
In multitask learning models, the feature generator is the same as fusion-based models but the classifier is composed of a bilinear upsampling layer that enlarges the feature map eight times and three convolution layers following \cite{kendall2017multi}.
We used Momentum SGD to optimize our model and set the momentum rate to $0.9$ and the learning rate to $1.0 \times 10^{−3}$ in all experiments. The image size was resized to $640 \times 480$ and no data augmentation methods were used.
We set one epoch to consist of 5000 iterations and chose test epoch numbers based on the entropy criteria following ~\cite{morerio2017minimal}.

\begin{figure}[t]
  \centering
\includegraphics[width=0.5\hsize]{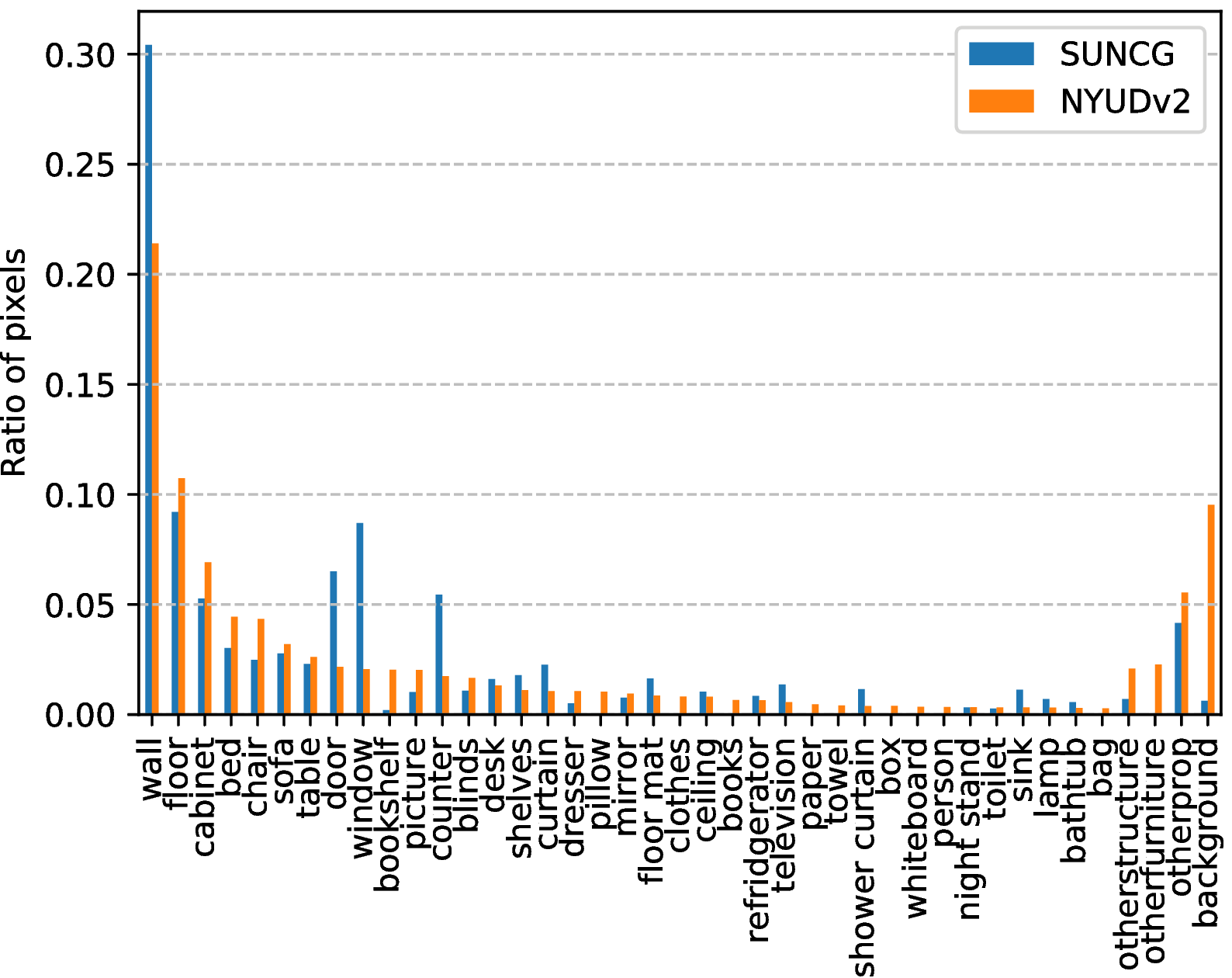}
\caption{Distribution of SUNCG~\cite{zhang2017physically} and NYUDv2~\cite{silberman2012nyu}.}
\label{fig:distribution}
\end{figure}

\noindent \textbf{Dataset:}
We used the publicly available synthetic dataset SUNCG~\cite{zhang2017physically} as the source domain dataset and the real dataset NYUDv2~\cite{silberman2012nyu} as the target domain dataset.
SUNCG contains two types of RGB images, an OpenGL-based and physically-based color image.
We would like to use more realistic data and therefore used the latter type.
568,793 RGB + HHA + instance boundary (only for multitask:triple) images of SUNCG and 795 RGB + HHA images of NYUDv2 train set were used for training and the NYUDv2 test set that contains 654 images was used for evaluation.
During training, we randomly sampled just a single sample (setting the batch size to 1 because of the GPU memory limit) from both the images (and their labels) of the source dataset and the remaining images of the target dataset yet with no labels.
Removing 6 classes (\textit{books}, \textit{paper}, \textit{towel}, \textit{box}, \textit{person}, \textit{bag}) that do not exist in SUNCG from the NYUDv2 40 class, 34 common classes were considered. According to the author of SUNCG, they removed \textit{person} and \textit{plant} in the rendered data because these two types of objects can be hardly rendered photo-realistic.
\figref{fig:distribution} shows the class label distribution of SUNCG and NYUDv2.
The imbalanced distribution demands the application of the four evaluation metrics.

\noindent \textbf{Evaluation Metrics:} We report on four metrics from common semantic segmentation and scene parsing evaluations. They are pixel accuracy (pixAcc), mean accuracy (mAcc), mean intersection over union (mIoU), and frequency weighted intersection over union (fwIoU). Let $k$ be the number of classes, $n_{ii}$ be the number of pixels of class $i$ predicted to belong to class $j$, $t_i$ be the total number of pixels of class $i$ in ground truth segmentation. We compute:
${\rm pixAcc} = \frac{\sum_{i} n_{ii}}{\sum_{i} t_{i}} (= \frac{\sum_{i} n_{ii}} {\sum_{i}\sum_{j} n_{ij}})$,
${\rm mAcc}   = \frac{1}{k} \sum_{i} \frac{n_{ii}}{t_{i}}$,
${\rm mIoU}   = \frac{1}{k} \sum_{i} \frac{n_{ii}}{\sum_{j} (n_{ij} + n_{ji})- n_{ii}}$,
${\rm fwIoU}  = \frac{1}{\sum_{k} t_{k}} \sum_{i} \frac{t_{i} n_{ii}}{\sum_{j} (n_{ij} + n_{ji})- n_{ii}}$.




\begin{figure*}[t]
  \centering
  \includegraphics[width=0.75\linewidth]{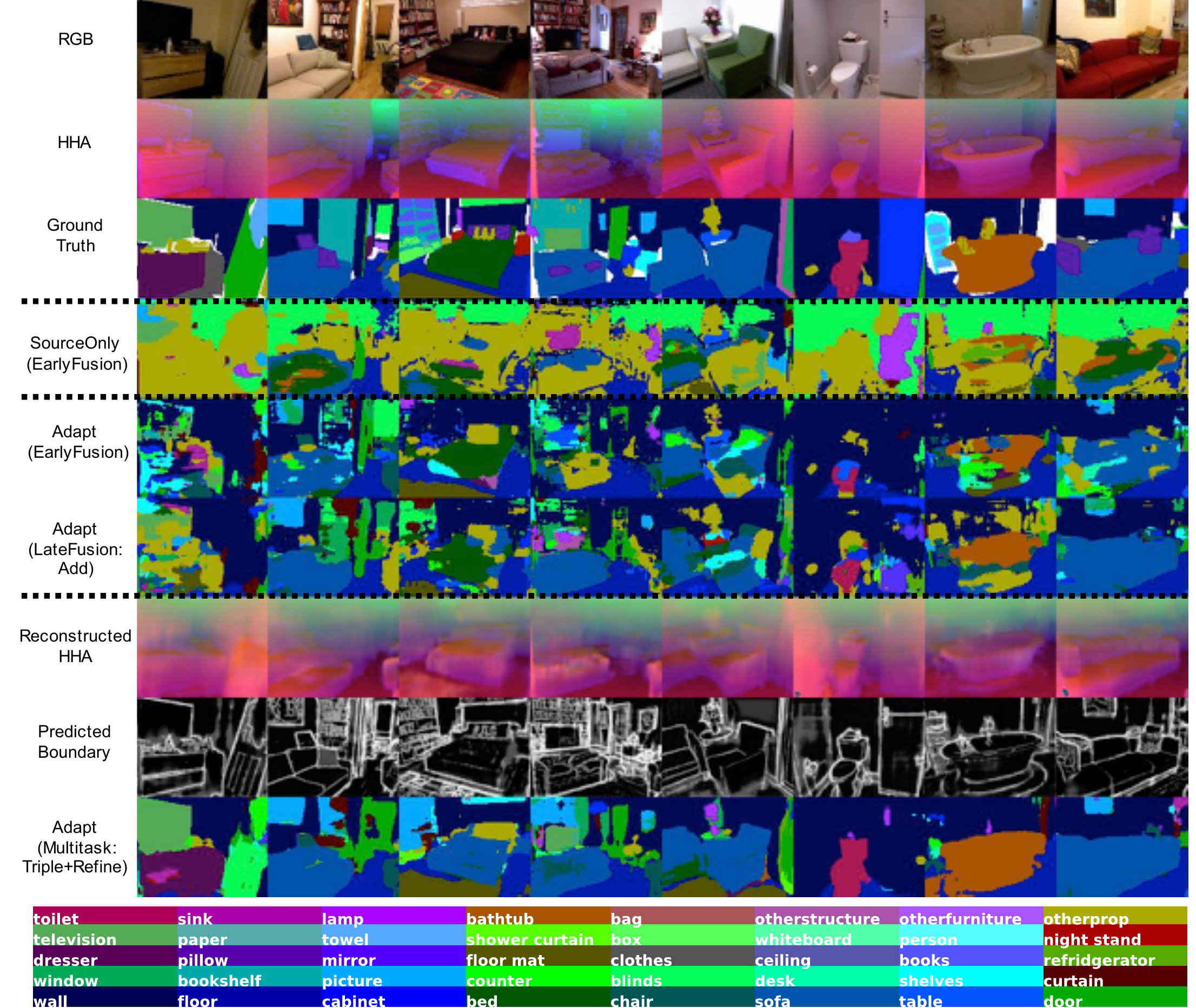} 
  \caption{Qualitative results. The three bottom rows show the results of \textit{Adapt (Multitask:Triple+Refine)}.}
  \label{fig:good_results}
\end{figure*}

\begin{table}[t]
\centering
\caption{Four evaluation metrics~[\%] of the the domain adaptation results from SUNCG~\cite{zhang2017physically} to NYUDv2~\cite{silberman2012nyu}. (Oracle is the result of the model trained on the train split of NYUDv2)}
\label{tab:nyu_total_result}
\scalebox{0.8} {
\begin{tabular}{l|rrrr}
\toprule[1.5pt]
                              &pixAcc& mAcc & fwIoU & mIoU \\ \hline \hline
Oracle (Target Only)           & 60.7 & 38.7 & 45.7 & 28.0 \\ \hline
Source Only (RGB)              & 13.0 & 6.7  & 9.9  & 3.2  \\
Source Only (HHA)              & 15.6   & 9.7  & 9.0   & 3.8  \\
Source Only (EarlyFusion)           & \textbf{17.9}   & \textbf{9.9}  & \textbf{10.0}  & \textbf{4.2}  \\ \hline
Adapt (RGB)                    & 42.3   & 19.1 & 27.2  & 11.4 \\
Adapt (HHA)                    & 40.5   & 13.4 & 22.7  & 8.6  \\
Adapt (EarlyFusion)            & \textbf{43.6}   & 17.1 & 27.6  & 10.7 \\
Adapt (LateFusion:Add)         & 41.8   & \textbf{19.7} & \textbf{28.6}  & \textbf{12.2} \\
Adapt (LateFusion:Concat)      & 40.9   & 17.1 & 25.8  & 10.6 \\
Adapt (ScoreFusion:Add)        & 39.7   & 19.0 & 27.6  & 10.7 \\
Adapt (ScoreFusion:ConcatConv) & 38.8   & 17.0 & 26.2  & 10.7 \\
Adapt (ScoreFusion:Gate)       & 37.3   & 14.8 & 24.1  & 9.2  \\
Adapt (FusenetFusion)        & 29.5   & 14.0 & 19.5  & 6.9  \\ \hline
Adapt (Multitask:Dual)          & \textbf{44.0}   & 20.2 & 30.1  & 12.8 \\
Adapt (Multitask:Triple)      & 42.6   & 22.6 & 30.0  & 13.1 \\
Adapt (Multitask:Triple+Refine) &  43.7   & \textbf{22.8} & \textbf{30.6}  & \textbf{13.2} \\
\bottomrule[1.5pt]
\end{tabular}
}
\end{table}

\begin{figure}[t]
  \centering
  \includegraphics[width=0.7\hsize]{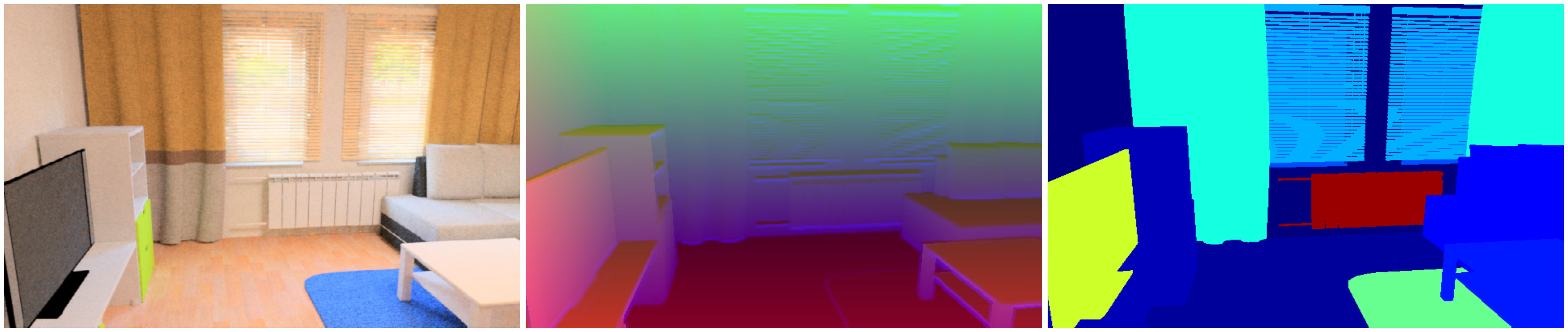}
  \caption{Sample (RGB, HHA, GT from left to right) of SUNCG~\cite{zhang2017physically}. \textit{Window}, \textit{Blinds} or \textit{Television} looks clear in the RGB image. However, \textit{Floor} looks clear in the HHA image.}
  \label{fig:tv_blinds_window}
\end{figure}


\begin{figure}[t]
  \centering
  \includegraphics[width=0.6\linewidth]{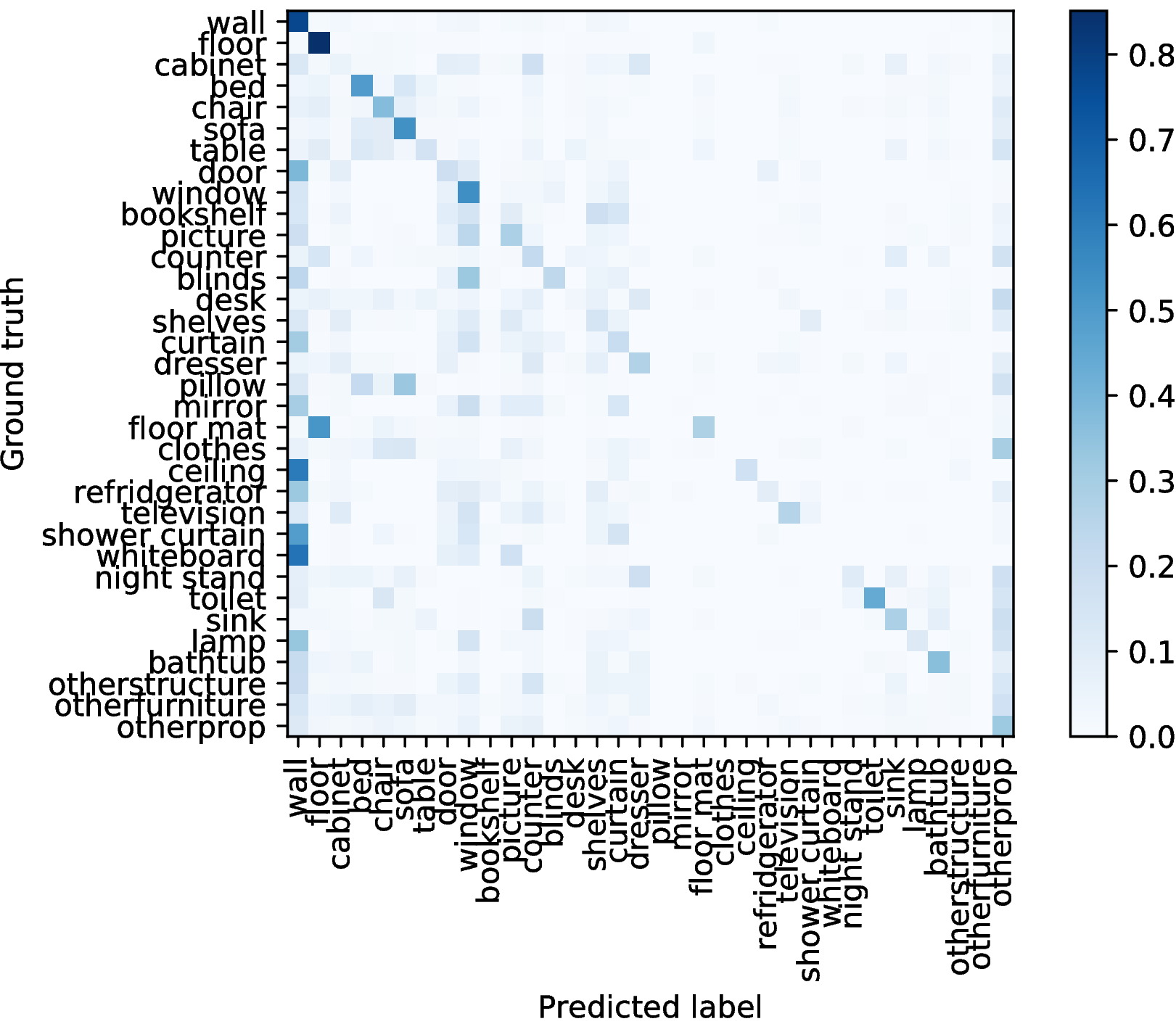}
  \caption{Confusion matrix of the result of \textit{Adapt (Multitask:Triple+Refined)}. Total value of each row is $1.0$.}
  \label{fig:conf_mat}
\end{figure}

\begin{figure}
\begin{minipage}{0.5\textwidth}
\begin{center}
\includegraphics[width=\hsize]{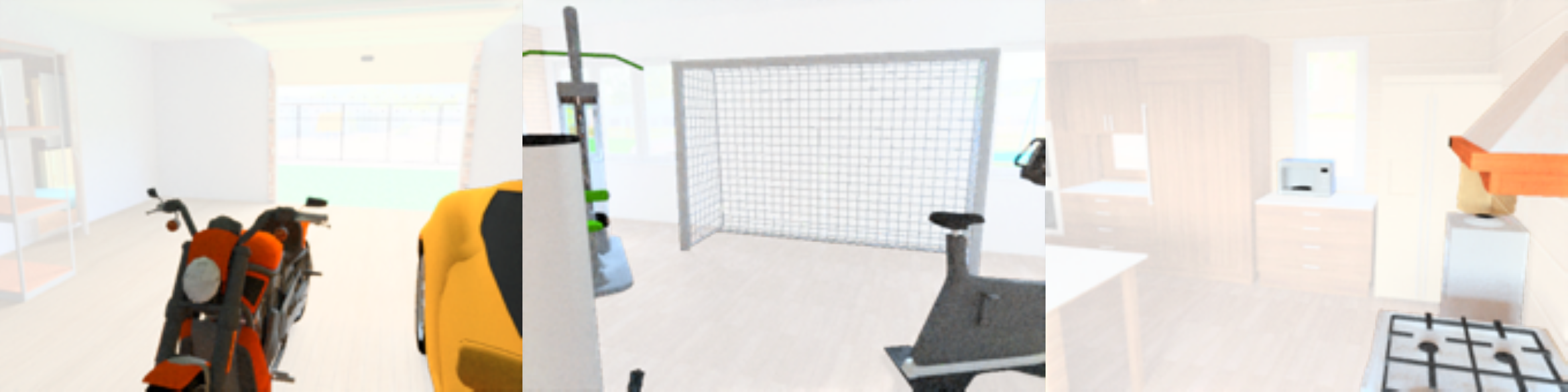}
\caption{Sample of \textit{Otherprop}.}
\label{fig:otherprop}
\end{center}
\end{minipage}
\begin{minipage}{0.5\textwidth}
\begin{center}
\makeatletter
\def\@captype{table}
\makeatother
\scalebox{0.8} {
\begin{tabular}{l|lll}
\toprule[1.5pt]
Method      &  ODS  & OIS   & AP    \\ \hline \hline
Sobel       & 0.446 & 0.473 & 0.190 \\
Canny       & 0.322 & 0.322 & 0.000 \\
Laplacian   & 0.450 & 0.467 & 0.262 \\ \hline
Source Only (Multitask:Triple) & 0.542 & 0.557 & \textbf{0.451} \\
Adapt (Multitask:Triple)       & \textbf{0.567} & \textbf{0.591} & 0.445 \\
\bottomrule[1.5pt]
\end{tabular}
}
\caption{Result of boundary detection．}
\label{tab:edge_detection_result}
\end{center}
\end{minipage}
\end{figure}

\begin{table}[t]
  \centering
  \caption{IoUs~[\%]  of the the domain adaptation results from SUNCG~\cite{zhang2017physically} to NYUDv2~\cite{silberman2012nyu}. (Oracle is the result of the model using train split of NYUDv2.)}
  \label{tab:nyu_iou}
\scalebox{0.68} { 
\begin{tabular}{l|rrrrrrrrrrrrrrrrrrrr}
  \toprule[1.5pt]
                              & \rotatebox{90}{Wall} & \rotatebox{90}{Floor} & \rotatebox{90}{Cabinet} & \rotatebox{90}{Bed} & \rotatebox{90}{Chair} & \rotatebox{90}{Sofa} & \rotatebox{90}{Table} & \rotatebox{90}{Door} & \rotatebox{90}{Window} & \rotatebox{90}{BookShelf} & \rotatebox{90}{Picture} & \rotatebox{90}{Counter} & \rotatebox{90}{Blinds} & \rotatebox{90}{Desks} & \rotatebox{90}{Shelves} & \rotatebox{90}{Curtain} & \rotatebox{90}{Dresser}  \\ \hline \hline
Oracle (Target Only)            & 66.9 & 78.8 & 43.3 & 53.6 & 39.7 & 38.6 & 25.8 & 16.3 & 33.9 & 32.1 & 43.6 & 38.9 & 42.5 & 9.2 & 7.0 & 24.0 & 19.0\\ \hline
Source Only (RGB)               & \textbf{27.0} & 19.5 & 1.4  & 0.2  & 1.7  & 2.3  & 3.9  & \textbf{8.5}  & \textbf{2.8}  & \textbf{0.1}  & 2.9  & 0.6  & 2.2  & 0.2 & 0.2 & 0.6  & \textbf{2.2}   \\
Source Only (HHA)               & 9.0  & 30.7 & \textbf{1.7}  & \textbf{11.5} & \textbf{13.3} & \textbf{8.7}  & 4.2  & 0.2  & 0.7  & 0.0  & 0.0  & 0.6  & 3.9  & 0.0 & \textbf{1.2} & 0.4  & 0.9  \\
Source Only (EarlyFusion)       & 7.5  & \textbf{46.8} & 0.2  & 11.0 & 3.0  & 8.1  & \textbf{6.1}  & 3.6  & 1.5  & 0.0  & \textbf{3.6}  & \textbf{2.4}  & \textbf{5.9}  & \textbf{0.4}  & 0.6 & \textbf{1.0}  & 0.7    \\ \hline
Adapt (RGB)                     & \textbf{54.4} & 56.8 & 5.8  & 12.7 & 27.8 & 28.8 & 10.6 & 8.7  & 13.3 & 0.5  & 9.0  & 4.1  & \textbf{26.3} & 4.1 & 2.0 & 9.4  & \textbf{8.8}  \\
Adapt (HHA)                     & 36.6 & \textbf{75.9} & 1.9  & 16.6 & 20.0 & 21.3 & 17.7 & 3.9  & 0.6  & 0.3  & 0.1  & 4.3  & 1.0  & \textbf{4.6} & 2.3 & 1.3  & 1.7   \\ %
Adapt (EarlyFusion)             & 50.1 & 75.8 & 2.6  & 19.4 & 24.7 & 19.8 & \textbf{21.3} & 6.1  & 8.3  & 0.4  & 6.0  & 3.5  & 8.9  & 2.9 & \textbf{4.0} & 5.1  & 2.7   \\ %
Adapt (LateFusion:Add)          & 50.4 & 74.3 & \textbf{6.7}  & 15.1 & \textbf{30.6} & \textbf{31.0} & 17.7 & 7.9  & 8.1  & 0.5  & 5.5  & 7.3  & 14.9 & 3.4 & 1.8 & 5.4  & 3.9    \\%
Adapt (LateFusion:Concat)       & 51.5 & 57.1 & 4.7  & 10.8 & 28.0 & 29.2 & 12.6 & \textbf{9.2}  & 7.9  & 0.2  & 9.3  & 2.2  & 17.4 & 3.2 & 2.0 & 3.1  & 5.6    \\ %
Adapt (ScoreFusion:Add)         & 50.9 & 74.3 & \textbf{6.7}  & 8.1  & 25.3 & 25.5 & 20.7 & 8.8  & 7.5  & \textbf{0.6}  & \textbf{9.9}  & 4.8  & 18.7 & \textbf{4.6} & 2.6 & 4.9  & 1.9  \\ %
Adapt (ScoreFusion:ConcatConv)  & 46.5 & 69.5 & 5.5  & \textbf{24.1} & 20.2 & 18.6 & 12.4 & 9.1  & 13.5 & 0.1  & 7.6  & \textbf{6.8}  & 13.7 & 3.6 & 1.3 & \textbf{11.4} & 2.2  \\ %
Adapt (ScoreFusion:Gate)        & 41.8 & 75.3 & 2.7  & 5.6  & 18.9 & 22.1 & 17.0 & 4.3  & \textbf{18.3} & 0.1  & 2.2  & 4.8  & 2.1  & 2.9 & 2.1 & 4.9  & 3.0    \\%
Adapt (FusenetFusion)           & 28.9 & 70.5 & 5.0  & 6.9  & 17.8 & 19.9 & 12.9 & 6.7  & 6.0  & 0.0  & 0.0  & 2.5  & 0.0  & 4.2 & 3.3 & 1.5  & 0.0   \\ \hline %
Adapt (Multitask:Dual)          & 55.5 & 67.6 & \textbf{7.8}  & 27.0 & \textbf{27.0} & 30.6 & \textbf{16.6} & \textbf{9.2}  & \textbf{18.2} & 0.0  & 16.0 & 7.4  & 15.1 & \textbf{2.7} & 3.0 & \textbf{10.6} & 2.6  \\ %
Adapt (Multitask:Triple)        & 55.3 & 67.6 & 4.2  & 33.0 & 25.9 & 31.0 & 12.0 & 8.1  & 15.9 & \textbf{0.4}  & 16.1 & \textbf{7.7} & 17.3 & 2.1 & \textbf{3.5} & 6.5  & 10.3 \\ %
Adapt (Multitask:Triple+Refine) & \textbf{56.7} & \textbf{68.2} & 4.2  & \textbf{35.0} & 26.0 & \textbf{32.6} & 12.2 & 8.5  & 16.3 & \textbf{0.4}  & \textbf{16.3} & \textbf{7.7}  & \textbf{18.4} & 1.9 & 3.4 & 6.8  & \textbf{10.8} \\
\bottomrule[1.5pt]

                                & \rotatebox{90}{Pillow} & \rotatebox{90}{Mirror} & \rotatebox{90}{Floor-mat} & \rotatebox{90}{Clothes} & \rotatebox{90}{Ceiling} & \rotatebox{90}{Refrigerator} & \rotatebox{90}{Television} &  \rotatebox{90}{Shower-curtain} & \rotatebox{90}{Whiteboard}  & \rotatebox{90}{NightStand} & \rotatebox{90}{Toilet} & \rotatebox{90}{Sink} & \rotatebox{90}{Lamp} & \rotatebox{90}{Bathtub} & \rotatebox{90}{Other-structure} & \rotatebox{90}{Other-furniture} & \rotatebox{90}{Other-prop} \\ \hline \hline
Oracle (Target Only)             & 27.2 & 10.9 & 17.9 & 14.2 & 58.0 & 9.2 & 30.1 & 6.5 & 0.1 & 2.1 & 51.1 & 28.3 & 23.5 & 18.5 & 9.0 & 6.1 & 27.4 \\ \hline
Source Only (RGB)              & 0.0  & 0.0  & 1.7 & 0.0  & 4.9  & 0.0 & \textbf{12.1} & 0.0 & 0.0 & 1.3 & 0.3  & 0.0  & \textbf{10.5} & 0.0  & 0.0 & 0.0 & 0.4  \\
Source Only (HHA)              & \textbf{4.5}  & 0.0  & 0.0  & 0.0  & \textbf{14.7} & \textbf{0.1} & 0.0  & 0.0 & 0.0 & \textbf{1.5} & 0.0  & \textbf{2.8}  & 3.8  & 2.6  & 0.0 & 0.0 & \textbf{13.8} \\
Source Only (EarlyFusion)      & 0.0  & 0.0  & \textbf{7.6}    & 0.0  & 6.1  & 0.0 & 2.8  & 0.0 & 0.0 & 0.3 & \textbf{0.8}  & \textbf{2.8}  & 3.0  & \textbf{5.0}  & \textbf{0.3} & 0.0 & 10.4 \\ \hline
Adapt (RGB)                    & 0.0  & 0.5  & 7.6  & 0.0  & 18.0 & 2.7 & \textbf{13.7} & 0.7 & 0.0 & 1.7 & 9.6  & 9.1  & \textbf{17.2} & 5.8  & \textbf{2.0} & 0.0 & \textbf{17.1} \\
Adapt (HHA)                    & 0.0  & 0.0  & 0.4  & 0.0  & 42.6 & 0.1 & 1.1  & 0.9 & 0.0 & 3.2 & 4.1  & 5.6  & 10.6 & 5.6  & 0.1 & 0.0 & 6.8  \\
Adapt (EarlyFusion)            & 0.0  & 0.1  & \textbf{12.3} & 0.0  & 29.4 & 1.3 & 6.0  & 1.3 & 0.0 & 1.0 & 9.1  & 5.9  & 10.9 & 6.8  & 1.5 & 0.0 & 15.3 \\
Adapt (LateFusion:Add)         & 0.0  & 2.0  & 6.0 & 0.0  & \textbf{45.8} & \textbf{4.3} & 11.5 & \textbf{2.3} & 0.0 & 3.2 & \textbf{16.6} & 6.0  & 8.9  & \textbf{10.1} & 0.6 & 0.0 & 14.5 \\
Adapt (LateFusion:Concat)      & 0.0  & 1.0  & 8.1 & 0.0  & 44.2 & 1.4 & 1.3  & 0.5 & 0.0 & \textbf{4.6} & 8.5  & 9.4  & 6.7  & 7.1  & 0.8 & 0.0 & 11.7 \\
Adapt (ScoreFusion:Add)        & 0.0  & 0.6  & 3.0  & 0.0  & 27.3 & 2.4 & 7.6  & 0.9 & 0.0 & 4.3 & 10.0 & 4.0  & 5.1  & 9.9  & \textbf{2.0} & 0.0 & 12.0 \\
Adapt (ScoreFusion:ConcatConv) & 0.0  & 1.0  & 2.7  & 0.0  & 30.4 & 0.9 & 18.2 & 0.1 & 0.0 & 0.4 & 5.3  & \textbf{10.0} & 8.1  & 5.6  & 0.3 & 0.0 & 13.9 \\
Adapt (ScoreFusion:Gate)       & 0.0  & 0.4  & 2.0 & 0.0  & 39.0 & 0.6 & 10.9 & 1.2 & 0.0 & 2.0 & 4.1  & 1.5  & 8.3  & 2.6  & 0.2 & 0.0 & 11.4 \\
Adapt (FusenetFusion)          & 0.0  & 0.0  & 1.4  & 0.0  & 38.1 & 0.0 & 4.6  & 0.0 & 0.0 & 0.0 & 0.0  & 0.0  & 0.2  & 0.0  & 0.0 & 0.0 & 4.2  \\ \hline
Adapt (Multitask:Dual)          & 0.0  & 0.8  & 7.8  & 0.0  & \textbf{36.5} & 2.7 & \textbf{12.7} & 1.6 & 0.0 & 2.6 & 11.5 & 3.1  & \textbf{12.0} & 10.2 & \textbf{1.7} & 0.0 & 16.6 \\
Adapt (Multitask:Triple)        & 0.0  & \textbf{1.4}  & 13.9 & 0.0  & 20.4 & \textbf{4.7} & 10.7 & 0.0 & 0.0 & 4.5 & 30.0 & \textbf{6.2}  & 6.7  & 10.0 & \textbf{1.7} & 0.0 & \textbf{17.6} \\
Adapt (Multitask:Triple+Refine) & 0.0  & 0.4  & \textbf{14.7} & 0.0  & 15.8 & \textbf{4.7} & 11.2 & 0.0 & 0.0 & \textbf{5.5} & \textbf{30.2} & \textbf{6.2}  & 5.7  & \textbf{11.2} & 1.6 & 0.0 & 17.3 \\
\bottomrule[1.5pt]
\end{tabular}
}
\end{table}

\subsection{Results} \label{sec:results}
\figref{fig:good_results} and \tabref{tab:nyu_total_result} show the qualitative and quantitative results, respectively.
We can confirm the effect of domain adaptation. \textit{Adapt (Multitask:Triple+Refine)} was the best according to three evaluation metrics and outperformed \textit{Adapt (RGB)} and \textit{Adapt (EarlyFusion)} in all the evaluation metrics.
The post-process which refines the segmentation results using the boundary detection outputs could improve all the metrics and lead better qualitative results.
In the fusion-based models, \textit{Adapt (LateFusion:Add)} could outperform \textit{Adapt (RGB)} in three evaluation metrics but the performance of most of the fusion-based models is not that different nor worse than that of \textit{Adapt (RGB)}.
This is due to the fact that the visibility of RGB images was better than that of HHA images for almost all the classes in the dataset in addition to the fact that objects which exist far away from the camera cannot be seen in HHA images due to the depth sensor range limitation, which could have a negative effect in fusion-based approaches.

\subsubsection{IoUs (see \tabref{tab:nyu_iou})}
\textbf{RGB v.s. HHA}
 The classes whose IoU of RGB outperformed those of HHA were \textit{Ceiling} and \textit{Floor}, whose visibility is better in HHA than in RGB. Conversely, the classes whose IoU of HHA outperformed those of RGB were \textit{Window}, \textit{Blinds} or \textit{Television}. This result is reasonable from the perspective of such classes as shown in: \figref{fig:tv_blinds_window}.

\textbf{Fusion-based v.s. Multitask learning}
Multitask learning approaches, especially the \textit{Adapt (Triple+Refine)} model, outperformed fusion-based approaches significantly in classes such as \textit{Bed}, \textit{Picture}, \textit{Toilet} except for \textit{Ceiling}. This indicates that multitask learning approaches work well for \textit{object} classes while fusion-based approaches work well for \textit{region} classes.

\subsubsection{Boundary detection result}
Following HED~\cite{xie2015holistically}, we computed three evaluation metrics, fixed contour threshold (ODS), per-image best threshold (OIS) and average precision (AP) using public code\footnote{https://github.com/pdollar/edges}.
We compared with handcrafted edge detection methods (Sobel, Canny, and Laplacian, whose hyper-parameter was set to default of OpenCV) that do not use ground truth.
As shown in \tabref{tab:edge_detection_result}, the boundary detection output outperformed these handcrafted methods, but the adaptation was not so effective. Based on this, boundary detection is considered to be a domain agnostic task compared to semantic segmentation.

\subsubsection{Failures}
From \figref{fig:distribution} and \tabref{tab:nyu_iou}, the IoUs of the rare classes such as \textit{BookShelf}, \textit{Pillow}, \textit{Mirror}, \textit{Clothes}, \textit{BookShel}, \textit{Shower-curtain}, \textit{Whiteboard}, \textit{Other-furniture} were zero or almost zero. Rare classes in source samples seem to be difficult to recognize.
\figref{fig:conf_mat} shows the confusion matrix of the best model, \textit{Adapt (Multitask:Triple+Refine)}.
\textit{Floormat} is mispredicted as \textit{Floor}, \textit{Whiteboard} is mispredicted as \textit{Wall},  \textit{Picture} and \textit{Blinds} are mispredicted as \textit{Window}, \textit{Pillow} is mispredicted as \textit{Sofa}.
If we consider the source label distribution shown in \figref{fig:distribution}, a rare class is often mispredicted as a common class whose position is the same as the rare one.
The ratio of pixels that are mispredicted as \textit{Othrprop} was high. This is probably because \textit{Othrprop} contains various kinds of classes such as \textit{car}, \textit{motorcycle}, \textit{soccer goal post}, \textit{gas stove} as shown in \figref{fig:otherprop}.


\section{Conclusion}
We combined a multichannel semantic segmentation task with an unsupervised domain adaptation task and proposed two architectures (fusion-based and multitask learning). We demonstrated that the multitask learning approach outperforms the simple \textit{early fusion} approach in all the evaluation metrics.
In addition, we propose adding a boundary detection task in the multitask learning approach and using the detection result to refine the segmentation output. We qualitatively and quantitatively show this post-process is effective especially in the classes whose boundaries look clear.
However, the scores of the adaptation result are still poor when compared to \textit{oracle}. In future work, we would like to use a few labeled and many unlabeled target samples (semi-supervised setting) and improve the results.

\section{Acknowledgements}
The work was partially funded by the ImPACT Program of the Council for
Science, Technology, and Innovation (Cabinet Office, Government of
Japan).

\clearpage

\bibliographystyle{splncs04}
\bibliography{bibTeX/domain_adaptation,bibTeX/segmentation,bibTeX/da_seg,bibTex/depth,bibTeX/multichannel_da,bibTeX/edge_detection}
\end{document}